\documentclass{article} 
\usepackage{iclr2027_conference,times}

\usepackage{amsmath,amsfonts,bm}

\def\eqref#1{equation~\ref{#1}}
\def\Eqref#1{Equation~\ref{#1}}

\def\1{\bm{1}}

\DeclareMathAlphabet{\mathsfit}{\encodingdefault}{\sfdefault}{m}{sl}
\SetMathAlphabet{\mathsfit}{bold}{\encodingdefault}{\sfdefault}{bx}{n}

\usepackage{hyperref}
\usepackage{url}
\usepackage{enumitem}
\usepackage{verbatim}   
\usepackage{amsmath,amssymb}
\usepackage{booktabs}
\usepackage{multirow}
\usepackage{makecell}
\usepackage{graphicx}
\usepackage[table]{xcolor}
\usepackage{wrapfig}
\usepackage{algorithm}
\usepackage{algorithmic}
\usepackage{pifont}
\usepackage{xspace}

\newcommand{\methodname}{DocSEAM\xspace}
\newcommand{\methodfull}{Document Selective Enrichment via Anchored Masking}

\title{DocSEAM: Bidirectional Cross-Page Encoding for Multi-Page Document Understanding}

\author{%
\hspace{-2.5mm} Hoang-Nhat Nguyen $^{1}$\quad Chen Change Loy$^{1,2}\thanks{Corresponding author.}$
\vspace{0.7mm}\\
\hspace{-2.5mm} $^{1}$S-Lab, Nanyang Technological University\quad $^{2}$Synvo AI
\vspace{0.7mm}\\
\hspace{-1.5mm}{\tt\small \url{https://mossbee.github.io/projects/DocSEAM}}
}

\iclrfinalcopy 

\hypersetup{
  pdftitle={DocSEAM: Bidirectional Cross-Page Encoding for Multi-Page Document Understanding},
  pdfauthor={Hoang-Nhat Nguyen and Chen Change Loy}
}
\begin{document}
\raggedbottom

\maketitle

\begin{abstract}

Multi-page document vision-language models encode each page independently, creating a feature-level discontinuity at page boundaries: visual tokens are formed without access to content on adjacent pages, and the language model must reason over these page representations formed in isolation.
We introduce \methodname (\methodfull), a lightweight module inserted between a pretrained vision encoder and the language model that injects the inductive bias of local cross-page context into visual representations.
For each page, \methodname forms a local triplet with its two neighbors, applies center-anchored masked self-attention with directional salience gates to fuse information from neighbor pages, and returns only the enriched center-page tokens, leaving the visual token count unchanged.
After a supervised cross-page warm-up and multi-page fine-tuning with \methodname frozen, our 9B-parameter model achieves state-of-the-art results among the evaluated open-weight models of up to 10B parameters on five multi-page document VQA benchmarks.
We further show that under controlled re-pagination, \methodname reduces sensitivity to boundary placement and improves cross-page span localization.

\end{abstract}

\section{Introduction}

Multi-page document vision-language models (VLMs) \citep{qwen25vl, qwen3vl, qwen35, internvl3} typically encode each page independently and concatenate the resulting visual tokens into a single sequence for the language model.
This design is convenient and scalable, but it imposes an implicit assumption: the representation of a page can be fully determined without seeing its neighbors.
In practice, this assumption is routinely violated.
Paragraphs continue across page breaks, tables carry rows onto the next page, and section headings introduce content that begins on the following leaf.
When a visual encoder commits to a representation of the last row of a table \emph{before} seeing its continuation on the next page, the language model inherits a feature-level discontinuity that it must repair from already-formed tokens.

\begin{figure}[t]
\centering

\includegraphics[width=0.9\textwidth]{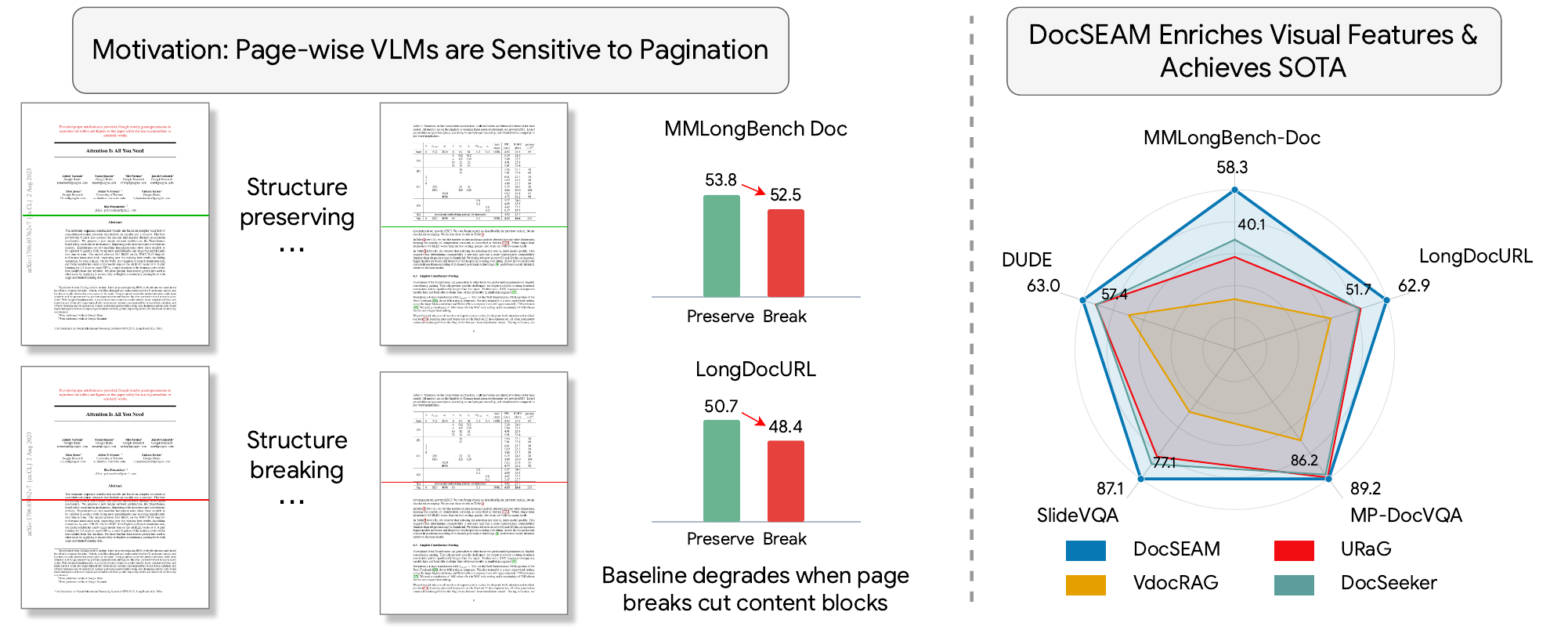}
\vskip -0.3 cm

\caption{
\textbf{Left:} Page-wise VLMs are sensitive to pagination.
Re-paginating the same documents with twice as many pages reduces baseline performance; breaking document structure by cutting through tables and paragraphs causes a further drop.
\methodname reduces the performance gap between structure-preserving and structure-breaking re-pagination to 0.8 and 0.7 percentage points on MMLongBench-Doc and LongDocURL, respectively (\S\ref{sec:pagination_invariance}).
\textbf{Right:} \methodname achieves the best results among the evaluated open-weight models of up to 10B parameters on five multi-page document VQA benchmarks.
}

\label{fig:teaser}
\end{figure}

Can the language model repair this discontinuity on its own?
In many cases, yes: the LLM can attend across pages, compare redundant evidence, and compensate for missing context.
But we show empirically that this compensation is fragile.
As shown in Fig.~\ref{fig:teaser}, when we re-paginate the same documents so that page breaks fall \emph{inside} tables and paragraphs rather than between them, page-wise VLMs suffer a consistent performance drop, even though the semantic content is unchanged (More analysis in Section~\ref{sec:pagination_invariance}).

This pagination sensitivity suggests that independently formed visual representations may contribute to the model's difficulty with cross-page reasoning: the language model must integrate information from page representations that were formed without access to neighboring pages.

We address this with a simple inductive bias: \emph{a page's visual representation should be informed by its immediate neighbors before reaching the language model.}
We propose \textbf{\methodname} (\textbf{Doc}ument \textbf{S}elective \textbf{E}nrichment via \textbf{A}nchored \textbf{M}asking), a lightweight module inserted between a pretrained VLM's vision encoder and its language model.
For each page~$p$, \methodname forms a local triplet $(p{-}1, p, p{+}1)$, applies a small number of self-attention layers over the concatenated triplet, and returns only the enriched center-page tokens.
The module adds ${\sim}$235M parameters (${\sim}$2.6\% of a 9B backbone), does not modify the vision encoder or change the visual token budget, and is initialized as a no-op so that training starts from the pretrained model's known-good operating point.

Two design choices encode document-specific structure.
First, a center-anchored attention mask allows the center page to attend to both neighbors while blocking direct interaction between the previous and next pages; this forces all cross-page information to flow \emph{through} the center page, preventing two-hop leakage between non-adjacent pages within a single layer.
Second, directional salience gates, consisting of four learnable gates per token, modulate every cross-page attention cell, allowing the model to learn \emph{where} cross-page communication is useful.

In a representative cross-page localization example, attention with
\methodname concentrates on boundary-spanning evidence at an earlier
LLM layer than with the baseline (Section~\ref{attention_trace}).
Our quantitative evaluation also shows improved cross-page span
localization accuracy, indicating that \methodname helps the model
identify evidence across page boundaries
(Section~\ref{sec:cross_page_analysis}).

Our \textbf{contributions} are threefold:
First, we identify pagination sensitivity as a symptom in conventional document VLMs and design a controlled re-pagination diagnostic to measure it.
Second, we propose \methodname, a simple post-encoder module that introduces the inductive bias of local cross-page context into multi-page document representations via center-anchored attention and directional salience gates.
Third, we design an effective two-stage training strategy in which \methodname is first warmed up on supervised cross-page alignment tasks and then \emph{frozen} during downstream fine-tuning, allowing the learned cross-page prior to transfer seamlessly across diverse tasks.
Together, these contributions enable \methodname to achieve state-of-the-art performance across five representative multi-page document VQA benchmarks (Fig.~\ref{fig:teaser}) among end-to-end open-weight VLMs with fewer than 10B parameters, a practically relevant model scale for document understanding.

\section{Related Work}
\label{sec:related_work}

\noindent\textbf{Multi-page Document Visual Question Answering.}
Document VQA was originally posed as a single-page problem~\citep{docvqa}.
Subsequent benchmarks extended it to multi-page settings: MP-DocVQA~\citep{mpdocvqa}, DUDE~\citep{dude}.
Current approaches fall into three families.
\emph{Retrieval-augmented methods} such as VisRAG~\citep{visrag}, SV-RAG~\citep{svrag} and VDocRAG~\citep{vdocrag} select relevant pages before reading them, but still encode each page independently.
\emph{Agentic methods} like MACT~\citep{mact} decompose reasoning into multi-step workflows (analysis, localization, reasoning) or multi-agent pipelines, yet the per-page visual representation is never modified.
\emph{End-to-end methods} such as GRAM~\citep{gram}, DocSeeker~\citep{docseeker}, mPLUG-DocOwl2~\citep{mplugdocowl2}, URaG~\citep{urag}, and Leopard~\citep{leopard} compress or reallocate visual tokens but still treat each page as an independently-encoded unit. 
Across all three families, a visual token near a page break is computed as if the page were self-contained; whatever cross-page reasoning happens, it happens \emph{after} that representational commitment.
\methodname modifies the page representation \emph{before} it reaches the language model, retriever, or agent, and does so densely (every token) and directionally (tied to the specific neighbor).

\noindent\textbf{Visual Encoders for Documents.}
A parallel line of work improves how a single page is converted into visual tokens.
Native-resolution ViTs~\citep{patchnpack} avoid aspect-ratio distortion on dense text;
DAVE~\citep{dave} pretrains an encoder specifically for documents and web UIs;
DeepEncoder~V2~\citep{deepseekocr2} learns intra-page reading order via causal queries;
UNIT~\citep{unit} unifies image and text recognition within one vision encoder and mPLUG-DocOwl2's H-Reducer compresses tokens while preserving layout.
All of these improve \emph{intra-page} representation quality.
\methodname is complementary: it sits after whatever per-page encoder produced the token grid and opens a channel between grids that were previously computed in isolation. 

\noindent\textbf{Stateful and Multi-image Encoding.}
Most multi-image VLMs~\citep{mantis,llavanextinterleave,llavaonevision} use late fusion: images are encoded independently and cross-image reasoning is deferred to the LLM. Streaming VLMs propagate visual memory forward~\citep{videollamb,videostreaming,flashvstream}, but this memory is causal and mainly targets temporal compression. SVE~\citep{sve} is the closest architectural prior: it injects causal cross-image interaction inside the ViT stack so the current image can condition on a previous image. \methodname{} is complementary rather than a direct variant. It leaves the vision encoder unchanged, operates post-encoder on the token grid consumed by the LLM, and uses a bidirectional three-page neighborhood because documents are fully observed at inference. Its gates are also document-specific, learning which boundary tokens should send or receive cross-page context while suppressing margins, headers, and page numbers. Appendix~\ref{sec:appendix_sve} compares SVE and \methodname under document fine-tuning.
\section{Method}
\label{sec:method}

\begin{figure}[t]
\centering
\includegraphics[width=\textwidth]{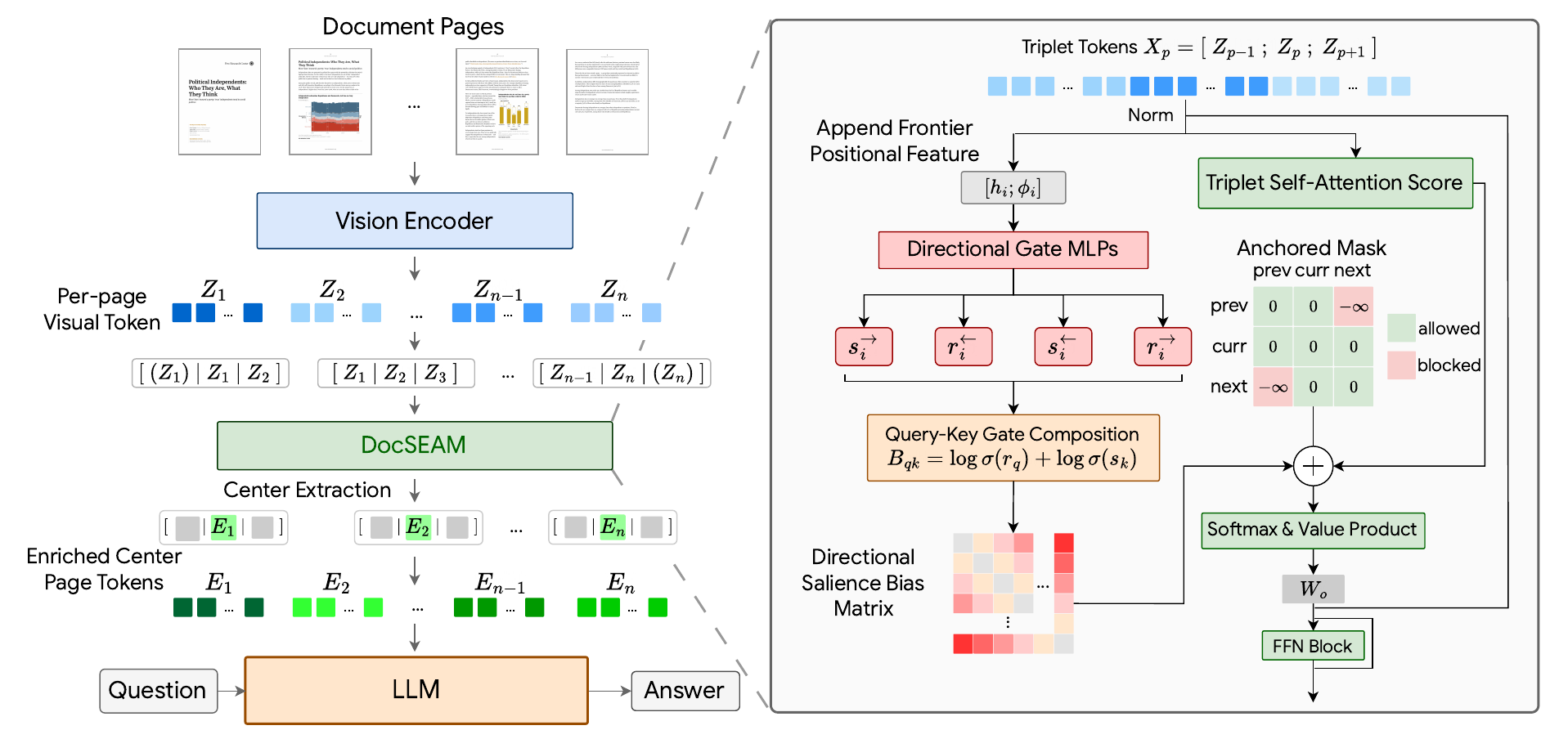}
\caption{\textbf{\methodname{} overview.}
For each page $p$, \methodname{} forms a local triplet $(Z_{p-1}, Z_p, Z_{p+1})$, applies center-anchored self-attention with directional salience gates, and returns only the enriched center-page tokens $E_p$. The language model receives the same number of visual tokens as in the stateless baseline.}
\label{fig:architecture}
\end{figure}

Let a document consist of $P$ page images $\{I_1,\ldots,I_P\}$.
A pretrained VLM vision encoder produces one visual-token sequence per page:
\begin{equation}
Z_p \in \mathbb{R}^{N_p \times d},
\label{eq:page_tokens}
\end{equation}
where $N_p$ is the number of visual tokens for page~$p$ and $d$ is the hidden dimension consumed by the language model. With native-resolution preprocessing, $N_p$ depends on the page resolution and aspect ratio after resizing and merging, and can therefore vary across pages.
In the standard multi-page pipeline, the sequences $Z_1,\ldots,Z_P$ are concatenated and fed directly into the language model.
Each page's representation is formed in complete isolation: the visual tokens at the bottom of page~$p$ are produced without access to any pixel on page~$p{+}1$, and vice versa.

\methodname replaces each $Z_p$ with an \emph{enriched} sequence of the same shape:
\begin{equation}
E_p = \mathrm{DocSEAM}(Z_{p-1},\, Z_p,\, Z_{p+1}) \;\in\; \mathbb{R}^{N_p \times d}.
\label{eq:tcea_output}
\end{equation}
The final visual-token sequence is $[E_1;\ldots;E_P]$, so the language model receives exactly the same number of visual tokens as in the baseline.
\methodname is inserted after the vision encoder and before the language model's multimodal input assembly.
This placement has three practical advantages: it avoids modifying the pretrained vision tower, operates directly on the features the language model will consume, and makes the DocSEAM\,/\,no-DocSEAM comparison a single-site architectural difference.

\subsection{Document-Local Triplets}
\label{sec:triplets}

As shown in Fig.~\ref{fig:architecture}, for each page~$p$, \methodname forms a local triplet by concatenating the token grids of the previous, current, and next page:
\begin{equation}
X_p = [\,Z_{p-1} ;\; Z_p ;\; Z_{p+1}\,] \in \mathbb{R}^{(N_{p-1}+N_p+N_{p+1})\times d}.
\label{eq:triplet}
\end{equation}
At document boundaries we use \emph{self-fallback}, where $Z_0 = Z_1$, and $Z_{P+1} = Z_P$, keeping the computation shape uniform without introducing a learned null-page embedding.

Each \methodname layer then applies masked self-attention over the triplet with two additive terms that control cross-page information flow:
\begin{equation}
\mathrm{Attn}(Q,K,V) = \mathrm{softmax}\!\left(\frac{QK^\top}{\sqrt{d_h}} + M + B\right)V,
\label{eq:attention}
\end{equation}
where $M$ is a center-anchored triplet mask that determines \emph{which} page groups may attend to each other (Section~\ref{sec:asymmetric_attention}), and $B$ is a directional salience bias that modulates \emph{how strongly} individual tokens participate in cross-page attention (Section~\ref{sec:gates}).
We describe each component below.

\subsection{Center-Anchored Triplet Attention}
\label{sec:asymmetric_attention}

Each \methodname layer performs self-attention over the concatenated triplet $X_p$.
Tokens from all three pages act as queries, keys, and values, but a center-anchored mask (Fig.~\ref{fig:architecture}) restricts which page groups may communicate.
Let the three groups in the triplet be \textsc{prev}, \textsc{curr}, and \textsc{next}.
The allowed attention flow is:
\begin{equation}
\begin{array}{rcl}
\textsc{prev} &\to& \{\textsc{prev},\, \textsc{curr}\}, \\[2pt]
\textsc{curr} &\to& \{\textsc{prev},\, \textsc{curr},\, \textsc{next}\}, \\[2pt]
\textsc{next} &\to& \{\textsc{curr},\, \textsc{next}\}.
\end{array}
\label{eq:mask}
\end{equation}
Direct $\textsc{prev} \leftrightarrow \textsc{next}$ interaction is blocked.
The two neighbor pages are separated by two page transitions; they should exchange information only through the center-page representation.
With $L$ stacked layers, the center page can still integrate information from neighbor states that interacted with it in earlier layers.

As shown in Fig.~\ref{fig:architecture}, a \methodname layer consists of layer normalization, a fused QKV projection, scaled dot-product attention with the center-anchored additive mask, an output projection, and a feed-forward residual block.
The output projection and the second feed-forward projection are \emph{zero-initialized}, so at initialization the residual branch contributes zero and \methodname is an exact identity, $E_p = Z_p$.

This property ensures that training begins from the pretrained model's original behavior; the cross-page contribution is learned as a residual on top of a known-good operating point.

\subsection{Directional Salience Gates}
\label{sec:gates}

Tokens near page boundaries are not equally informative: headers, footers, page numbers, and blank margins often lie near edges but rarely provide continuation evidence.
\methodname{} therefore learns \emph{directional salience gates} that selectively modulate cross-page attention cells.

For each token~$i$ with hidden representation~$h_i$, the module computes four scalar logits using small independent MLPs:
\begin{equation}
s_i^{\to},\quad r_i^{\gets},\quad s_i^{\gets},\quad r_i^{\to},
\label{eq:gate_logits}
\end{equation}
where $s^{\to}$ and $s^{\gets}$ denote sending information to the next and previous pages, respectively, while $r^{\gets}$ and $r^{\to}$ denote receiving information from the previous and next pages.

Each send logit is produced by a two-layer MLP that receives both the layer-normalized visual token representation and a sinusoidal positional embedding $\phi_i \in \mathbb{R}^{18}$ encoding token distance to the active content frontiers:
\begin{equation}
s_i^{(\cdot)} = f_\theta^{(\cdot)}\!\bigl([\mathrm{LN}(h_i);\, \phi_i]\bigr).
\label{eq:gate_mlp}
\end{equation}

The receive logits are computed in the same way, using their own independent MLPs. The positional features use content frontiers as reference points,
making them less sensitive to varying page margins. Details of the positional encoding and gate-input ablations are given
in Appendices~\ref{sec:gate_details} and~\ref{sec:appendix_gate_inputs}.

The gates enter attention as additive log-biases before the softmax.
For a current-page query~$q$ reading a previous-page key~$k$:
\begin{equation}
B_{qk} = \log\sigma(r_q^{\gets}) + \log\sigma(s_k^{\to}).
\label{eq:bias_prev}
\end{equation}
Here, $\sigma(r_q^{\gets})$ controls the current-page token's reception from the previous page, while $\sigma(s_k^{\to})$ controls the previous-page token's transmission to the next page.
The two gates therefore describe the receiving and sending ends of the same previous-to-current-page link.
For a current-page query reading a next-page key:
\begin{equation}
B_{qk} = \log\sigma(r_q^{\to}) + \log\sigma(s_k^{\gets}).
\label{eq:bias_next}
\end{equation}
The same directional convention applies when neighbor pages query center-page keys.
Same-page attention cells receive no salience bias ($B_{qk}=0$).
Substituting $M$ and $B$ into \Eqref{eq:attention} completes the \methodname attention layer.
Since the exponentiated bias factorizes as
\begin{equation}
\exp(B_{qk}) = \sigma(r_q)\,\sigma(s_k),
\label{eq:gate_product}
\end{equation}
cross-page communication is strong only when both the receiver's demand and the sender's utility are jointly high.
We refer to this as the \emph{dual-agreement} property: a cross-page link is activated only when both sides of the boundary agree that the interaction is useful.
When either gate is near zero, $B_{qk} \to -\infty$ and the corresponding attention weight is suppressed regardless of the QK similarity.

\subsection{Training}
\label{sec:training}

Training proceeds in two stages.

\noindent\textbf{Stage A: Cross-page Boundary Alignment.} \label{sec:stage_a} Stage~A trains only \methodname while keeping the vision tower and language model frozen. 
We draw supervised examples from MP-DocStruct1M~\citep{mplugdocowl2}, pairing consecutive-page tasks (requiring cross-page evidence) with single-page controls (preserving single-page reading).

The Stage~A objective combines next-token cross-entropy loss with a lightweight gate alignment loss:
\begin{equation}
\mathcal{L}_{A} = \mathcal{L}_{\mathrm{CE}} + \lambda_{\mathrm{gate}} \mathcal{L}_{\mathrm{gate}},
\label{eq:stage_a_loss}
\end{equation}
where $\mathcal{L}_{\mathrm{gate}} = \mathcal{L}_{\mathrm{frontier}} + \lambda_{\mathrm{margin}}\mathcal{L}_{\mathrm{margin}} + \mathcal{L}_{\mathrm{budget}}$ supplies weak geometric guidance. The frontier term aligns directional send/receive gates with the corresponding ink boundaries, the margin term suppresses activations over whitespace, and $\mathcal{L}_{\mathrm{budget}}=\max(0,\bar g-\rho)^2$ penalizes mean gate activation above $\rho=0.20$. Cross-entropy supplies the task-level supervision; further details are in Appendix~\ref{sec:gate_details}.
Self-supervised boundary reconstruction was discarded as it led to gate collapse: global self-attention within each page allows masked tokens to be reconstructed without cross-page context (Appendix~\ref{sec:appendix_representation_leakage}).

\noindent\textbf{Stage B: Multi-page Document Fine-tuning.} \label{sec:stage_b} In Stage~B, \methodname is \emph{frozen} and the rest of the model is fine-tuned with LoRA~\citep{lora} on a mixture of multi-page document understanding tasks.
Freezing \methodname preserves its parameters learned in Stage~A while the downstream model adapts to the enriched tokens. The stateless baseline uses the same training-data pool and LoRA configuration, with Stage~A and Stage~B data combined in one SFT stage; training-strategy controls are reported in Section~\ref{sec:ablations}.
Full dataset composition and hyperparameters for both stages are provided in Appendix~\ref{sec:training_details}.

\noindent\textbf{Inference.} At inference, each page is encoded once by the vision tower.
\methodname processes the page triplets in a single sliding-window pass, producing enriched tokens at no increase in the visual sequence length seen by the language model. Because the enrichment depends only on document pages, the enriched tokens can be computed once and reused across multiple questions about the same document.
The additional inference cost of \methodname is small: two transformer layers over triplets of post-merger tokens, amounting to less than 7\% of the total forward-pass FLOPs for a typical 20-page document. 

\section{Experiments}
\label{sec:experiments}

\noindent\textbf{Backbone.}
We instantiate \methodname on Qwen3.5-9B~\citep{qwen35}, a native-resolution VLM with a NaViT-style vision tower and a $2{\times}2$ pixel-shuffle merger.
\methodname adds $L{=}2$ transformer layers (${\sim}235$M parameters) between the Vision Encoder and the 9B language model.
To demonstrate generality, we also report results with Qwen3.5-4B, Gemma3-4B and Qwen2.5-VL-7B backbones. 

\noindent\textbf{Benchmarks.}
We evaluate on five multi-page document VQA benchmarks:
DUDE~\citep{dude}, MP-DocVQA~\citep{mpdocvqa}, SlideVQA~\citep{slidevqa},
MMLongBench-Doc~\citep{mmlongbenchdoc}, and LongDocURL~\citep{longdocurl}.
DUDE and MP-DocVQA use ANLS, while SlideVQA, MMLongBench-Doc, and LongDocURL use accuracy.
We also report cross-page parsing results on MPDocBench-Parse~\citep{mpdocbenchparse}.
Dataset details and task characteristics are provided in Appendix~\ref{app:benchmarks}.

\noindent\textbf{Baselines.}
Our primary comparison is a \emph{matched stateless baseline}: the same Qwen3.5-9B backbone, the same LoRA configuration, trained on the combined Stage~A and Stage~B data in a single SFT stage. We show in Section~\ref{sec:ablations} that adding Stage~A data to the baseline yields negligible change, so this setup is strictly more favorable to the baseline than using Stage~B data alone. We additionally compare against other multi-page document understanding methods.

\subsection{Main Results}
\label{sec:main_results}

We compare \methodname against the proprietary models GPT-4o~\citep{gpt4o} and Gemini 3 Pro~\citep{gemini3}, as well as open-weight methods M3DocRAG~\citep{m3docrag}, DocOwl2~\citep{mplugdocowl2}, Vis-RAG~\citep{visrag}, SV-RAG~\citep{svrag}, VDocRAG~\citep{vdocrag}, Docopilot~\citep{docopilot}, InternVL3~\citep{internvl3}, DocSeeker~\citep{docseeker}, and URaG~\citep{urag}.
We also include the zero-shot Qwen3.5~\citep{qwen35} backbone and its matched SFT baseline.

\begin{table}[t]
\caption{Multi-page document VQA results.
All models in the open-weight block use comparable backbone sizes ($\leq$10B).
DUDE and MP-DocVQA are measured by ANLS; SlideVQA, MMLongBench-Doc, and LongDocURL by accuracy.
$\dagger$: proprietary model. Best open-weight result in \textbf{bold}.}
\vspace{1ex}
\label{tab:main_results}
\centering
\setlength{\tabcolsep}{4pt}
\begin{tabular}{lccccccc}
\toprule
Method & Venue & Size & DUDE & \makecell{MP-\\DocVQA} & \makecell{Slide\\VQA} & \makecell{MMLong\\Bench} & \makecell{Long\\DocURL} \\
\midrule
\multicolumn{8}{l}{\textit{Proprietary}} \\
GPT-4o$^\dagger$ & -- & -- & 54.1 & 67.4 & -- & 42.8 & 64.5 \\
Gemini 3 Pro$^\dagger$ & -- & -- & 70.1 & -- & -- & 59.5 & 72.0 \\
\midrule
\multicolumn{8}{l}{\textit{Open-weight ($\leq$10B)}} \\
M3DocRAG & arXiv'24 & 9B & -- & 84.4 & 55.7 & 21.0 & 35.1\\
DocOwl2 & ACL'25 & 8B & 46.8 & 69.4 & -- & 13.4 & 5.3 \\
Vis-RAG & ICLR'25 & 4B & -- & 70.9 & 50.7 & 18.8 & 41.9 \\
SV-RAG & ICLR'25 & 4B & 45.0 & 71.0 & 23.0 & -- & -- \\
VDocRAG & CVPR'25 & 8B & 44.0 & 62.6 & 42.0 & 18.4 & 39.8 \\
Docopilot & CVPR'25 & 8B & -- & 81.3 & -- & 28.8 & -- \\
InternVL3 & arXiv'25 & 8B & 47.4 & 80.8 & 38.7 & 24.1 & 54.4 \\
DocSeeker & CVPR'26 & 7B & 57.4 & 86.2 & 77.1 & 40.1 & 51.7 \\
URaG & AAAI'26 & 7B & 57.6 & 88.2 & 72.1 & 33.8 & 52.2 \\
\midrule
Qwen3.5 & -- & 9B & 46.8 & 84.2 & 76.6 & 57.1 & 54.5 \\
Qwen3.5 + SFT & -- & 9B & 54.1 & 87.5 & 85.1 & 44.5 & 42.8 \\
\methodname + Qwen2.5 VL & -- & 7B & 58.0 & 88.3 & 80.0 & 39.5 & 53.3 \\
\methodname + Qwen3.5 & -- & 4B & 57.5 & 86.3 & 79.6 & 38.2 & 53.3 \\
\textbf{\methodname + Qwen3.5} & -- & 9B & \textbf{63.0} & \textbf{89.2} & \textbf{87.1} & \textbf{58.3} & \textbf{62.9} \\
\bottomrule
\end{tabular}
\end{table}

Table~\ref{tab:main_results} presents the main results.
\methodname + Qwen3.5-9B achieves the best performance among the evaluated open-weight models of up to 10B parameters on all five benchmarks: 63.0 ANLS on DUDE (+5.4 over URaG), 89.2 ANLS on MP-DocVQA (+1.0 over URaG), 87.1 accuracy on SlideVQA (+10.0 over DocSeeker), 58.3 accuracy on MMLongBench-Doc (surpassing even Qwen3.5 zero-shot and all other open-weight models), and 62.9 accuracy on LongDocURL (+10.7 over URaG).

Notably, standard SFT on short-answer document QA data actually \emph{degrades} performance on MMLongBench-Doc ($57.1 \to 44.5$) and LongDocURL ($54.5 \to 42.8$) compared to the Qwen3.5 zero-shot baseline, likely because fine-tuning on short, extractive answers biases the model away from the multi-step reasoning these benchmarks require.
\methodname not only reverses this degradation but surpasses the zero-shot baseline on both benchmarks (58.3 and 62.9), demonstrating that cross-page enrichment provides complementary benefits that survive task-specific fine-tuning.
The gains over the matched SFT baseline are substantial across all benchmarks: +8.9 on DUDE, +1.7 on MP-DocVQA, +2.0 on SlideVQA, +13.8 on MMLongBench-Doc, and +20.1 on LongDocURL.

\methodname is backbone-agnostic.
When applied to Qwen2.5-VL-7B, the same backbone used by DocSeeker and URaG, \methodname achieves 58.0 on DUDE and 88.3 on MP-DocVQA, surpassing both published methods with a general-purpose module rather than task-specific training.
Even the smaller \methodname + Qwen3.5-4B variant reaches 57.5 on DUDE and 86.3 on MP-DocVQA, performing on par with DocSeeker (57.4, 86.2) and URaG (57.6, 88.2) despite using roughly half the backbone parameters. Full zero-shot and SFT comparisons for the 4B and 7B backbones are in Table~\ref{tab:smaller_backbones} in the appendix. We further validate \methodname on cross-page document parsing (MPDocBench-Parse) and stratified evidence-topology analysis (LongDocURL) in Appendix~\ref{sec:appendix_parsing} and~\ref{sec:appendix_stratified}, respectively, confirming that gains generalize beyond VQA and concentrate on adjacent-page evidence.

\subsection{Cross-Page Representation Analysis}
\label{sec:cross_page_analysis}

To understand \emph{how} \methodname improves downstream performance, we examine the cross-page representations through attention visualization, a large-scale localization benchmark, and gate activation patterns.

\begin{figure}[tb]
\centering
\includegraphics[width=\textwidth]{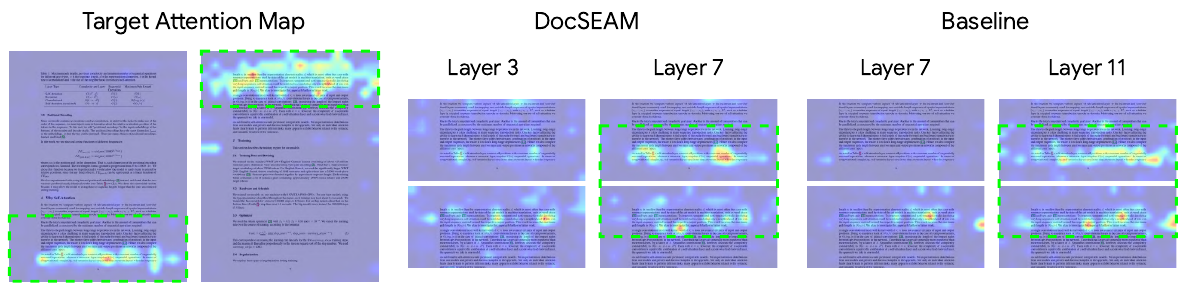}
\vskip -0.25cm
\caption{LLM layer-wise attention trace for cross-page span localization. Given a text span query crossing a page boundary, \methodname concentrates attention on the ground-truth evidence in early LLM layers (Layer 7), whereas the baseline forms a comparable pattern only in late layers (Layer 11). For readability, layer heatmaps display cropped top/bottom boundary regions of consecutive pages. Green and red dashed boxes indicate correct and missing attention, respectively.}
\label{fig:span_attention_trace_first}
\end{figure}

\noindent\textbf{Attention trace: early-layer cross-page localization.}
\label{attention_trace}
We visualize the layer-wise attention maps of the language model when queried with a cross-page span localization prompt.
Figure~\ref{fig:span_attention_trace_first} shows a representative case: the model is asked which pages contain a quoted text span that crosses a page boundary.
With \methodname, attention concentrates on the correct boundary-spanning evidence as early as Layer~7, whereas the baseline model without cross-page enrichment fails to form a comparable pattern until Layer~11.
This earlier focus suggests that \methodname makes boundary-spanning evidence more readily accessible to the language model.
An extended visualization comparing attention maps across deeper LLM layers is provided in Appendix~\ref{sec:appendix_extended_attention_trace}.

\noindent\textbf{Large-scale span localization benchmark.}
To evaluate output localization beyond individual attention examples, we construct a quantitative benchmark of 1{,}500 samples from MPDocBench-Parse~\citep{mpdocbenchparse} with three task types (500 each):
\emph{cross-page text lookup} (identify \emph{both} pages of a boundary-spanning span),
\emph{single-page text lookup} (identify the page of a page-local span), and
\emph{single-page existence} (judge whether a span appears on a given page).

\begin{table}[tb]
\caption{Cross-page span localization diagnostic.
Exact-match accuracy (\%) on three probe types.
The cross-page lookup requires identifying \emph{both} pages of a boundary-spanning span.}
\label{tab:span_localization}
\begin{center}
\small
\begin{tabular}{lccc}
\toprule
Model & \makecell{Cross-page\\lookup} & \makecell{Single-page\\lookup} & \makecell{Existence\\check} \\
\midrule
Qwen3.5 (zero-shot) & 50.2 & 86.0 & 91.8 \\
Qwen3.5 + SFT (baseline) & 65.0 & 70.0 & \textbf{95.6} \\
\textbf{Qwen3.5 + \methodname (ours)} & \textbf{85.8} & \textbf{94.2} & 94.0 \\
\bottomrule
\end{tabular}
\end{center}
\end{table}

Table~\ref{tab:span_localization} shows that \methodname achieves 85.8\% on cross-page lookup (+20.8 over the SFT baseline).
The SFT baseline's single-page lookup drops to 70.0\% compared to the zero-shot 86.0\%, likely due to distribution overfitting; \methodname recovers and surpasses it (94.2\%).
Existence checking, which requires no cross-page linking, is slightly lower with \methodname than with SFT (94.0\% vs.\ 95.6\%). These results show that \methodname improves the model's ability to locate both cross-page and single-page evidence, while retaining strong performance on the simpler existence check.

\begin{figure}[tb]
\centering
\includegraphics[width=0.9\textwidth]{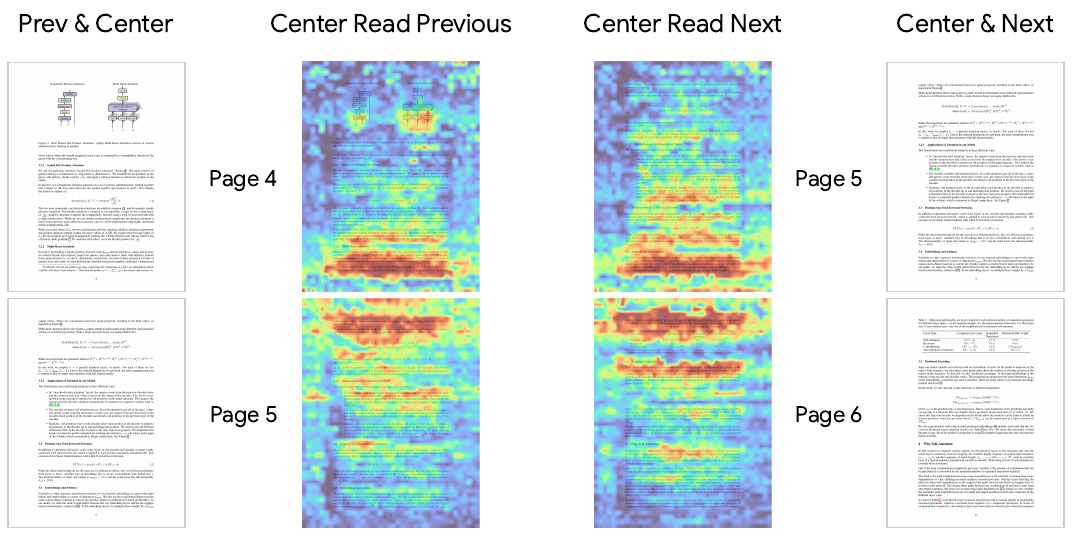}
\caption{\textbf{Cross-page attention within \methodname.}
Across a triplet from \emph{Attention Is All You Need} (Pages 4--6).
\textbf{(a--b)} Page~5 reads backward: attention flows to continuous text at the bottom of Page~4 and grounds on Figure~2.
\textbf{(c--d)} Page~5 reads forward: despite Table~1 intervening at the top of Page~6, \methodname attends to both the table and the text continuation beneath it.
Blank margins receive near-zero attention throughout.}
\label{fig:gate_heatmaps}
\end{figure}

\noindent\textbf{Gate activation patterns.}
Figure~\ref{fig:gate_heatmaps} visualizes the cross-page attention flow within the \methodname module across three consecutive pages of \emph{Attention Is All You Need} \citep{transformers}.
Three emergent behaviors stand out.
First, activation concentrates on the bottom and top of consecutive pages, where truncated content resides, while blank margins, headers, and page numbers receive near-zero attention in this example, consistent with selective routing.
Second, in the previous-page interaction (Panels a, b), the Multi-Head Attention figure on Page~4 lights up together with the corresponding equation on Page~5, consistent with a link between a diagram and its mathematical definition across the page boundary.
Third, in the next-page interaction (Panels c, d), Page~6 exhibits two distinct bright regions: Section~3.5 continues from the bottom of Page~5, but Table~1 is inserted between the boundary and the continuation; attention is elevated on both the table and the text below it, consistent with routing across the layout interruption.
Additional visualizations are in Appendix~\ref{sec:additional_visualizations}.

\subsection{Pagination Robustness}
\label{sec:pagination_invariance}

A model with robust document understanding should be insensitive to \emph{where} page breaks fall.
We test this by splitting every page of a benchmark into two half-pages near the center in two modes:
\textbf{structure-preserving} (cut at whitespace between paragraphs/tables) and \textbf{structure-breaking} (cut through paragraphs or table interiors).
Both variants have identical page counts and pixel content; only the boundary location differs.

\begin{table}[t]
\begin{minipage}[t]{0.48\textwidth}
\centering
\caption{\textbf{Pagination robustness} under controlled re-pagination (accuracy \%). $\Delta_{\mathrm{orig}} = S_{\mathrm{orig}} - S_{\mathrm{break}}$ measures total drop; $\Delta_{\mathrm{pag}} = S_{\mathrm{pres}} - S_{\mathrm{break}}$ isolates boundary-placement sensitivity.}
\label{tab:pagination_invariance}
\resizebox{\linewidth}{!}{%
\begin{tabular}{llcc}
\toprule
Metric & Model & \makecell{MMLong\\Bench} & \makecell{Long\\DocURL} \\
\midrule
\multirow{3}{*}{Original ($S_{\mathrm{orig}}$)} 
& Zero-shot baseline & 57.1 & 54.5 \\
& SFT baseline & 44.5 & 42.8 \\
& \textbf{+\methodname} & \textbf{58.3} & \textbf{62.9} \\
\midrule
\multirow{3}{*}{Preserved ($S_{\mathrm{pres}}$)} 
& Zero-shot baseline & 53.8 & 50.7 \\
& SFT baseline & 42.1 & 41.5 \\
& \textbf{+\methodname} & \textbf{57.6} & \textbf{61.8} \\
\midrule
\multirow{3}{*}{Broken ($S_{\mathrm{break}}$)} 
& Zero-shot baseline & 52.5 & 48.4 \\
& SFT baseline & 40.9 & 39.9 \\
& \textbf{+\methodname} & \textbf{56.8} & \textbf{61.1} \\
\midrule
\multirow{3}{*}{Degradation ($\Delta_{\mathrm{orig}}$)} 
& Zero-shot baseline & 4.6 & 6.1 \\
& SFT baseline & 3.6 & 2.9 \\
& \textbf{+\methodname} & \textbf{1.5} & \textbf{1.8} \\
\midrule
\multirow{3}{*}{Penalty ($\Delta_{\mathrm{pag}}$)} 
& Zero-shot baseline & 1.3 & 2.3 \\
& SFT baseline & 1.2 & 1.6 \\
& \textbf{+\methodname} & \textbf{0.8} & \textbf{0.7} \\
\bottomrule
\end{tabular}%
}
\end{minipage}%
\hfill
\begin{minipage}[t]{0.49\textwidth}
\centering
\caption{\textbf{Key ablations} on MMLongBench-Doc and LongDocURL (accuracy \%). $^\dagger$Baseline SFT trained on Stage~A+B data.}
\label{tab:ablations}
\resizebox{\linewidth}{!}{%
\begin{tabular}{lcc}
\toprule
Variant & \makecell{MMLong\\Bench} & \makecell{Long\\DocURL} \\
\midrule
Qwen3.5 + SFT (Stage B only) & 44.3 & 42.6 \\
Qwen3.5 + SFT$^\dagger$ (Stage A+B data) & 44.5 & 42.8 \\
\midrule
\multicolumn{3}{l}{\textit{Capacity controls}} \\
\;\; + Self-as-neighbor & 45.1 & 43.5 \\
\;\; + Dup.\ vision-enc.\ layers & 45.8 & 44.1 \\
\;\; + Single-page transformer & 46.2 & 44.5 \\
\midrule
\multicolumn{3}{l}{\textit{Module design}} \\
\;\; + Causal (prev-only) triplet & 53.1 & 55.8 \\
\;\; + Full 3-page attn., no gates & 55.2 & 57.8 \\
\;\; + Anchored mask, no gates & 55.9 & 58.7 \\
\;\; + Full 3-page attn.\ + 4 dir.\ gates & 56.8 & 60.5 \\
\;\; + Anchored mask + 4 dir.\ gates (ours) & \textbf{58.3} & \textbf{62.9} \\
\midrule
\multicolumn{3}{l}{\textit{Training strategy}} \\
\;\; Baseline two-stage (no \methodname) & 42.1 & 40.3 \\
\;\; No Stage A (direct Stage B) & 52.7 & 56.1 \\
\;\; Stage A: downstream mix + aux. & 53.9 & 57.6 \\
\;\; Stage A: boundary data, CE only & 55.8 & 59.5 \\
\;\; Stage A: boundary data + aux.\ (ours) & \textbf{58.3} & \textbf{62.9} \\
\bottomrule
\end{tabular}%
}
\end{minipage}
\end{table}

We define two metrics to quantify robustness.
The \emph{total degradation} $\Delta_{\mathrm{orig}} = S_{\mathrm{orig}} - S_{\mathrm{break}}$ measures how much performance drops from the original to the worst-case re-pagination.
The \emph{pagination penalty} $\Delta_{\mathrm{pag}} = S_{\mathrm{preserve}} - S_{\mathrm{break}}$ isolates sensitivity to boundary placement by comparing two re-paginated variants with identical page counts.

Table~\ref{tab:pagination_invariance} presents the comparison between \methodname and its matched SFT baseline (both trained on identical SFT context mix), alongside the zero-shot baseline. 
The matched Baseline SFT suffers a total degradation $\Delta_{\mathrm{orig}}$ of 3.6 points on MMLongBench-Doc and 2.9 points on LongDocURL, with pagination penalties $\Delta_{\mathrm{pag}}$ of 1.2 and 1.6 respectively. 
By injecting local cross-page context, \methodname dramatically reduces total degradation to 1.5 on MMLongBench-Doc and 1.8 on LongDocURL, while suppressing the pagination penalty to just 0.8 and 0.7. 
Under structure-breaking re-pagination, \methodname exceeds the original SFT scores on both benchmarks; it also exceeds the original zero-shot score on LongDocURL (61.1 vs.\ 54.5), while remaining slightly below it on MMLongBench-Doc (56.8 vs.\ 57.1). 
These results show improved robustness under the tested re-pagination protocol. A baseline trained with re-paginated data is reported in Appendix~\ref{sec:appendix_repagination_augmentation}.
The full re-pagination protocol, including the cutting algorithm and sample counts, is described in Appendix~\ref{sec:appendix_repagination}.

\subsection{Ablation Studies}
\label{sec:ablations}

Table~\ref{tab:ablations} presents key ablations on the two benchmarks most sensitive to cross-page reasoning.
\textbf{Capacity controls:} the ``self-as-neighbor'' variant uses the identical \methodname architecture and FLOPs but feeds copies of the center page instead of real neighbors; its small gain (+0.6 on MMLongBench) supports the role of real neighboring-page context beyond added capacity.
Two further controls match \methodname's parameter count without cross-page information: duplicating the last layers of the vision encoder and adding single-page transformer layers; both show similarly small gains (Appendix~\ref{sec:appendix_full_ablations}). \textbf{Bidirectional vs.\ causal:} restricting to a previous-page-only window degrades performance, validating that documents benefit from seeing both neighbors.
\textbf{Data attribution:} to ensure gains are architectural rather than data-driven, our SFT baseline is trained on the \emph{combined} Stage~A and Stage~B data in a single stage; adding Stage~A data yields negligible change over Stage~B-only SFT (44.5 vs.\ 44.3 on MMLongBench), suggesting that adding cross-page parsing data to standard SFT alone is insufficient to reproduce \methodname's gains.
\textbf{Training strategy:} applying the two-stage schedule to the stateless baseline reduces performance (42.1 vs.\ 44.5 on MMLongBench). To isolate the boundary curriculum from dedicated module adaptation, we instead train \methodname in Stage~A on the downstream mixture, keeping the same steps, batch size, CE and auxiliary losses, and Stage~B procedure. This control reaches 53.9 / 57.6, compared with 52.7 / 56.1 without Stage~A and 58.3 / 62.9 with boundary-focused Stage~A. The additional +4.4 / +5.3 points support the boundary curriculum beyond providing an adaptation stage. Self-supervised boundary reconstruction also reduces performance (Appendix~\ref{sec:appendix_full_ablations}).
Full ablations including gate input variants, number of \methodname layers, and backbone generality are in Appendix~\ref{sec:appendix_full_ablations}.

\section{Conclusion}

We identified pagination sensitivity as a weakness of page-wise document VLMs and proposed DocSEAM, a lightweight post-encoder module that enriches each page’s visual tokens with bidirectional neighbor context through center-anchored attention and directional salience gates, without modifying the vision encoder or increasing the visual token budget. 
DocSEAM achieves state-of-the-art results among the evaluated open-weight models ($\leq$10B) on five multi-page document VQA benchmarks, improves cross-page span localization by 20.8 points over the SFT baseline, and reduces the pagination penalty to 0.8 and 0.7 points on MMLongBench-Doc and LongDocURL, respectively. 
Ablations suggest that neighboring-page context contributes to the gains beyond added capacity, while gate analysis shows directional specialization near content boundaries. These results highlight the value of incorporating local page continuity into document representations.

\bibliography{iclr2027_conference}
\bibliographystyle{iclr2027_conference}

\appendix
\newpage

\section{Training Details}
\label{sec:training_details}

\paragraph{Stage A dataset.}
We derive 74k training examples from MP-DocStruct1M~\citep{mplugdocowl2}, organized into four groups:
(1)~\textbf{Consecutive parsing} (50k): the model reads text spanning the boundary between two consecutive pages and produces the merged passage.
(2)~\textbf{Consecutive lookup} (12k): given a text snippet crossing a page boundary, the model identifies which pages contain it.
(3)~\textbf{Single-page parsing control} (6k): the model reads a single page's text inside a multi-page context.
(4)~\textbf{Single-page lookup control} (6k): the model localizes page-local evidence inside a multi-page context.
All examples are presented as document-local page neighborhoods.
 
\paragraph{Stage A hyperparameters.}
$L{=}2$ \methodname layers, AdamW with learning rate $10^{-4}$, cosine decay, batch size 64 (8 GPUs $\times$ gradient accumulation 8), 500 training steps.

\paragraph{Stage B dataset.}
The training mixture includes multi-page document VQA datasets (MP-DocVQA~\citep{mpdocvqa}, DUDE~\citep{dude}, SlideVQA~\citep{slidevqa}) and MP-Docreason51K multipage reasoning data from mPlug-DocOwl2 ~\citep{mplugdocowl2}.

\paragraph{Stage B hyperparameters.}
LoRA~\citep{lora} rank 16 applied to all linear layers of the language model. Visual patch merger is also fine-tuned; vision tower is frozen.
AdamW with learning rate $2{\times}10^{-5}$, cosine schedule, batch size 64, 2 epochs.

\section{Gate and Loss Details}
\label{sec:gate_details}

\paragraph{Content-frontier positional encoding.}
The positional feature $\phi_i \in \mathbb{R}^{18}$ encodes each token's distance to the top and bottom content frontiers (first and last rows containing document ink).
The distance to each frontier is capped at a fraction $\alpha = 0.20$ of the page height and mapped via a 4-frequency sinusoidal expansion:
\begin{equation}
\psi(d) = \bigl[d,\, \sin(\pi d),\, \cos(\pi d),\, \ldots,\, \sin(8\pi d),\, \cos(8\pi d)\bigr] \in \mathbb{R}^9,
\end{equation}
yielding $\phi_i = [\psi(d_{\mathrm{top}}(r_i));\, \psi(d_{\mathrm{bot}}(r_i))]$.
Physical page edges often contain semantically irrelevant white space; measuring distance relative to detected ink frontiers rather than raw margins provides the gate with a geometric prior invariant to blank page margins.

\paragraph{Auxiliary gate loss.}
The gate alignment loss consists of three terms:
\begin{equation}
\mathcal{L}_{\mathrm{gate}} = \mathcal{L}_{\mathrm{frontier}} + \lambda_{\mathrm{margin}} \mathcal{L}_{\mathrm{margin}} + \mathcal{L}_{\mathrm{budget}},
\end{equation}
where $\mathcal{L}_{\mathrm{frontier}}$ applies a weighted binary cross-entropy to align directional send/receive channels with content frontiers, $\mathcal{L}_{\mathrm{margin}}$ suppresses gate leakage over whitespace margins, and $\mathcal{L}_{\mathrm{budget}} = \max(0, \bar{g} - \rho)^2$ penalizes mean gate activation above the target budget $\rho = 0.20$.

\section{Benchmark Details}
\label{app:benchmarks}

\paragraph{Document VQA benchmarks.}
We evaluate on five multi-page document VQA benchmarks that cover complementary document types and reasoning regimes.
\textbf{DUDE}~\citep{dude} is built from multi-industry, multi-domain, multi-page visually rich documents and includes diverse question and answer types, including layout-navigating and multi-step reasoning questions.
We report ANLS following the benchmark protocol.
\textbf{MP-DocVQA}~\citep{mpdocvqa} extends single-page document VQA to multi-page scanned industry documents by providing each question with its surrounding document pages, up to 20 pages per document.
We report ANLS.
\textbf{SlideVQA}~\citep{slidevqa} evaluates question answering over slide decks, where the model must identify relevant evidence slides and answer the question.
We report exact-match accuracy.
\textbf{MMLongBench-Doc}~\citep{mmlongbenchdoc} targets long-context document understanding over lengthy PDF documents with evidence drawn from text, tables, charts, images, and layout structure.
It includes both single-page and cross-page questions, as well as unanswerable questions for hallucination evaluation.
We report accuracy.
\textbf{LongDocURL}~\citep{longdocurl} evaluates long multimodal document understanding through tasks covering document understanding, numerical reasoning, and cross-element locating.
We report accuracy.

\paragraph{Parsing benchmark.}
Beyond VQA, we evaluate on \textbf{MPDocBench-Parse}~\citep{mpdocbenchparse}, a multi-page document parsing benchmark designed for document-level end-to-end parsing.
Its evaluation protocol covers content fidelity and logical structure, including text, table, and formula recognition, truncated text and table merging, figure extraction, reading order, and heading hierarchy recovery.
Because \methodname{} is designed to improve cross-page continuity, we focus on the two metrics most directly tied to page-boundary reasoning: truncated text merging and truncated table merging.

\section{LongDocURL Stratified Analysis}
\label{sec:appendix_stratified}

LongDocURL provides \texttt{evidence\_pages} annotations for each question, enabling us to stratify results by evidence topology without constructing new data.

\begin{table}[h]
\caption{LongDocURL stratified by evidence topology.
$\Delta_{\text{Base}}$: improvement of \methodname over the zero-shot baseline; $\Delta_{\text{SFT}}$: improvement of \methodname over Baseline SFT.}
\label{tab:stratified}
\begin{center}
\small
\begin{tabular}{lccccc}
\toprule
Evidence topology & Zero-shot & SFT & \methodname & $\Delta_{\text{Base}}$ & $\Delta_{\text{SFT}}$ \\
\midrule
Single page & 57.5 & 47.5 & 63.0 & +5.5 & +15.5 \\
Adjacent two pages & 58.0 & 47.0 & 69.5 & +11.5 & +22.5 \\
Multiple ($\geq$3, $\exists$ adj.) & 48.8 & 38.4 & 60.1 & +11.3 & +21.7 \\
Non-adjacent only & 48.8 & 40.0 & 54.7 & +5.9 & +14.7 \\
\midrule
Overall & 54.5 & 42.8 & 62.9 & +8.4 & +20.1 \\
\bottomrule
\end{tabular}
\end{center}
\end{table}

Across both baseline comparisons, the largest gains occur on \textbf{adjacent two-page} questions (+11.5 over zero-shot Baseline, +22.5 over Baseline SFT), precisely the setting where \methodname's triplet window directly enriches both evidence pages.
Multiple-page questions with at least one adjacent pair also benefit substantially (+11.3 over Baseline, +21.7 over SFT).
Single-page questions show moderate gains (+5.5 over Baseline, +15.5 over SFT), while non-adjacent questions improve modestly (+5.9 over Baseline, +14.7 over SFT), consistent with the local window providing less direct benefit for distant evidence.

Standard SFT without cross-page enrichment degraded performance across all evidence topologies compared to the zero-shot baseline (dropping overall from 54.5 to 42.8), likely because fine-tuning on short extractive answers biases the model away from long-context multi-step reasoning. 
In contrast, \methodname not only fully reverses this degradation across every category, but yields substantial improvements over SFT (most prominently on adjacent page splits ($\Delta_{\text{SFT}} = +22.5$)). 
The larger gains on adjacent-page evidence under both baseline comparisons are consistent with \methodname's local-context bias. This stratification alone does not establish the causal mechanism of the improvement.

\section{Multi-Page Document Parsing}
\label{sec:appendix_parsing}

To test whether \methodname's benefit extends beyond QA, we evaluate on MPDocBench-Parse~\citep{mpdocbenchparse}, a recent benchmark designed to assess multi-page document parsing.
We focus on the two metrics most directly related to cross-page continuity: \textbf{truncated text merging} (T-Text) and \textbf{truncated table merging} (T-Table), where existing models nearly universally fail.

\begin{table}[h]
\caption{MPDocBench-Parse results on cross-page parsing.
T-Text and T-Table measure merging of paragraphs and tables split across pages.
NED$\downarrow$ for T-Text; TEDS$\uparrow$ for T-Table.
$^*$Uses explicit rule-based merging heuristics.}
\label{tab:mpdocbench}
\begin{center}
\small
\begin{tabular}{lccc}
\toprule
Model & Overall$\uparrow$ & T-Text$\downarrow$ & T-Table$\uparrow$ \\
\midrule
\multicolumn{4}{l}{\textit{End-to-end specialized}} \\
FireRed-OCR & 69.3 & 0.178 & 62.6 \\
DeepSeek-OCR2 & 75.6 & 0.284 & 63.0 \\
dots.ocr & 74.2 & 0.303 & 60.7 \\
\midrule
\multicolumn{4}{l}{\textit{General VLMs}} \\
Gemini 3.1 Pro & 71.8 & 0.232 & 61.3 \\
Qwen3-VL-235B & 74.0 & 0.187 & 61.3 \\
InternVL3.5-38B & 57.0 & 0.515 & 51.7 \\
\midrule
\multicolumn{4}{l}{\textit{Pipeline-based$^*$}} \\
PaddleOCR-VL-1.5$^*$ & 80.8 & \textbf{0.147} & 83.1 \\
MinerU2.5$^*$ & 76.7 & 0.328 & \textbf{88.1} \\
\midrule
\multicolumn{4}{l}{\textit{Ours}} \\
Qwen3.5 + SFT (base.) & 55.2 & 0.341 & 55.8 \\
\textbf{Qwen3.5 + \methodname} & 65.6 & 0.172 & 62.9 \\
\bottomrule
\end{tabular}
\end{center}
\end{table}

Table~\ref{tab:mpdocbench} shows that \methodname substantially improves cross-page parsing.
On truncated text merging, \methodname reduces Normalized Edit Distance (NED) from 0.341 to 0.172, surpassing all end-to-end VLMs and approaching the best pipeline-based model (PaddleOCR-VL-1.5, 0.147) which employs explicit rule-based merging heuristics.
On truncated table merging, \methodname improves TEDS from 55.8 to 62.9, a +7.1 point gain over the baseline.

This cross-task validation is important: the same architectural module that improves QA also improves parsing, suggesting that \methodname learns a general property of cross-page document representation rather than a task-specific optimization.
The pipeline-based models achieve higher T-Table scores using hand-designed rules to detect and stitch truncated tables. Without explicit merging logic, \methodname improves over the baseline, though a substantial gap to these pipelines remains.

\section{Full Ablation Details}
\label{sec:appendix_full_ablations}

Table~\ref{tab:full_ablations} extends the main-paper ablations (Table~\ref{tab:ablations}) with additional variants.

\begin{table}[h]
\caption{Full ablation results on MMLongBench-Doc and LongDocURL (accuracy~\%).}
\label{tab:full_ablations}
\begin{center}
\small
\begin{tabular}{lcc}
\toprule
Variant & \makecell{MMLong\\Bench} & \makecell{Long\\DocURL} \\
\midrule
Qwen3.5 + SFT (no \methodname) & 44.5 & 42.8 \\
\midrule
\multicolumn{3}{l}{\textit{Capacity controls}} \\
\;\; + Self-as-neighbor & 45.1 & 43.5 \\
\;\; + Dup.\ vision-enc.\ layers & 45.8 & 44.1 \\
\;\; + Single-page transformer & 46.2 & 44.5 \\
\midrule
\multicolumn{3}{l}{\textit{Module design}} \\
\;\; + Causal (prev-only) triplet & 53.1 & 55.8 \\
\;\; + Full 3-page attn., no gates & 55.2 & 57.8 \\
\;\; + Anchored mask, no gates & 55.9 & 58.7 \\
\;\; + Full 3-page attn.\ + 4 dir.\ gates & 56.8 & 60.5 \\
\;\; + Anchored mask + shared gate & 57.1 & 60.2 \\
\;\; + Anchored mask + 4 dir.\ gates (ours) & \textbf{58.3} & \textbf{62.9} \\
\midrule
\multicolumn{3}{l}{\textit{Number of \methodname layers}} \\
\;\; 1 layer & 54.1 & 60.8 \\
\;\; 2 layers (ours) & \textbf{58.3} & \textbf{62.9} \\
\;\; 4 layers & 55.1 & 61.7 \\
\midrule
\multicolumn{3}{l}{\textit{Training strategy}} \\
\;\; Baseline two-stage (no \methodname) & 42.1 & 40.3 \\
\;\; No Stage A (direct Stage B) & 52.7 & 56.1 \\
\;\; Stage A: downstream mixture, CE + aux. & 53.9 & 57.6 \\
\;\; Stage A: boundary reconstruction & 47.3 & 48.9 \\
\;\; Stage A: boundary data, CE only & 55.8 & 59.5 \\
\;\; Stage A: boundary data, CE + aux.\ (ours) & \textbf{58.3} & \textbf{62.9} \\
\bottomrule
\end{tabular}
\end{center}
\end{table}

\paragraph{Capacity controls.}
All three controls match \methodname's ${\sim}$235M parameter budget without cross-page information.
\emph{Self-as-neighbor} uses the identical architecture but feeds copies of the center page as neighbors (+0.6 on MMLongBench).
\emph{Duplicated vision-encoder layers} extends the per-page visual encoder to the same depth.
\emph{Single-page transformer} adds independent transformer layers over each page's tokens.
None exceeds +1.7 over the baseline, supporting the benefit of real cross-page context beyond a comparable parameter increase.

\paragraph{Module design.}
Restricting \methodname to a causal (previous-only) window loses 5.2 points on MMLongBench, validating bidirectional enrichment.
The center-anchored mask consistently outperforms full 3-page attention under matched gating (58.3 vs.\ 56.8 with four gates; 55.9 vs.\ 55.2 without gates), supporting the design choice of blocking direct prev$\leftrightarrow$next interaction.
Four directional gates outperform a shared gate by 1.2 on MMLongBench and 2.7 on LongDocURL.

\paragraph{Depth.}
One \methodname layer already yields +9.6 over the baseline on MMLongBench; two layers provide the best result. Four layers slightly regress, likely due to over-smoothing of page-specific features.

\paragraph{Training strategy and data attribution.}
To ensure fair attribution, Table~\ref{tab:data_attribution} specifies the data and trainable components for each compared variant.

\begin{table}[h]
\caption{Training configuration for data-attribution controls. The downstream-mixture control and our model use the same Stage~A steps, batch size, CE and auxiliary gate losses, and Stage~B procedure.}
\label{tab:data_attribution}
\begin{center}
\small
\resizebox{\linewidth}{!}{%
\begin{tabular}{lccc}
\toprule
Variant & Stage A data & Stage B data & Trainable in Stage A \\
\midrule
SFT baseline (B only) & \ding{55} & \ding{51} & - \\
SFT baseline (A+B)$^\dagger$ & \multicolumn{2}{c}{combined, single stage} & LoRA \\
Baseline two-stage & \ding{51} & \ding{51} & LoRA (no \methodname) \\
\methodname (mixture control) & Downstream mixture & \ding{51} & \methodname only \\
\methodname (ours) & Boundary data & \ding{51} & \methodname only \\
\bottomrule
\end{tabular}%
}
\end{center}
\end{table}

The SFT baseline used throughout the paper trains on the combined Stage~A and Stage~B data in a single stage ($^\dagger$).
This gives the baseline access to the additional Stage~A data, yet yields nearly identical performance (44.5 vs.\ 44.3 on MMLongBench), suggesting that this data addition alone has little effect under the tested SFT configuration.
The \emph{baseline two-stage} variant applies the same two-stage schedule to the baseline without \methodname; performance decreases to 42.1 / 40.3 from 44.5 / 42.8.

To distinguish dedicated module adaptation from the boundary-focused curriculum, we train only \methodname in Stage~A on the full downstream training mixture. We keep the number of steps, batch size, CE plus auxiliary gate losses (including their weights), and subsequent Stage~B procedure identical to our boundary-alignment setup. This control reaches 53.9 / 57.6, improving over no Stage~A (52.7 / 56.1) by 1.2 / 1.5 points. Boundary-focused Stage~A reaches 58.3 / 62.9, adding a further 4.4 / 5.3 points under the matched adaptation budget and objectives. These results support the value of the boundary curriculum beyond providing a dedicated adaptation stage.

On boundary data, CE-only Stage~A reaches 55.8 / 59.5; adding auxiliary gate losses yields a further 2.5 / 3.4 points. Self-supervised boundary reconstruction performs worse (47.3 / 48.9), consistent with the same-page shortcut discussed in Section~\ref{sec:stage_a}.

\paragraph{Backbone generality.}

\begin{table}[h]
\caption{\methodname consistently improves over matched baselines across VLM backbones.}
\label{tab:backbones}
\begin{center}
\small
\begin{tabular}{lcccccc}
\toprule
\multirow{2}{*}{Backbone} & \multicolumn{3}{c}{LongDocURL} & \multicolumn{3}{c}{MMLongBench} \\
\cmidrule(lr){2-4}\cmidrule(lr){5-7}
 & Base & Base + SFT & +\methodname & Base & Base + SFT & +\methodname \\
\midrule
Qwen2.5-VL-7B & 37.8 & 42.7 & 53.3 & 25.4 & 28.8 & 39.5 \\
Qwen3.5-9B    & 54.5 & 42.8 & 62.9 & 57.1 & 44.5 & 58.3 \\
Gemma3-4B     & 34.7 &  39.5    &   42.6   &  20.1    &   25.2   & 33.6 \\
\bottomrule
\end{tabular}
\end{center}
\end{table}

\begin{table}[t]
\caption{Zero-shot, matched SFT, and \methodname results for smaller backbones. DUDE and MP-DocVQA use ANLS; the remaining benchmarks use accuracy.}
\label{tab:smaller_backbones}
\centering
\resizebox{\linewidth}{!}{%
\begin{tabular}{llccccc}
\toprule
Backbone & Variant & DUDE & MP-DocVQA & SlideVQA & MMLongBench-Doc & LongDocURL \\
\midrule
Qwen3.5-4B & Zero-shot & 43.6 & 79.2 & 67.3 & 51.9 & 51.4 \\
 & SFT & 51.1 & 83.6 & 81.2 & 37.9 & 38.8 \\
 & +\methodname & 57.5 & 86.3 & 79.6 & 38.2 & 53.3 \\
\midrule
Qwen2.5-VL-7B & Zero-shot & 35.2 & 70.1 & 59.8 & 25.4 & 37.8 \\
 & SFT & 56.0 & 82.9 & 67.4 & 28.8 & 42.7 \\
 & +\methodname & 58.0 & 88.3 & 80.0 & 39.5 & 53.3 \\
\bottomrule
\end{tabular}%
}
\end{table}

Table~\ref{tab:smaller_backbones} provides the full five-benchmark comparison. \methodname improves over SFT on all five benchmarks for Qwen2.5-VL-7B and four of five for Qwen3.5-4B. On the 4B backbone, SlideVQA drops from 81.2 to 79.6, and MMLongBench-Doc improves only slightly over SFT (37.9 to 38.2), remaining below zero-shot (51.9). LongDocURL improves over both SFT and zero-shot (53.3 versus 38.8 and 51.4). The benefit therefore varies across backbones and tasks. Our backbone experiments cover the 4B, 7B, and 9B model scales; larger backbones have not been evaluated.

\methodname yields consistent gains across three backbone families.
On Qwen2.5-VL-7B, the same backbone used by DocSeeker and URaG, \methodname surpasses both published methods with a general-purpose module.

\paragraph{Inference overhead.}
Each page undergoes one additional forward pass through the \methodname module (${\sim}$235M parameters) within a sliding-window triplet.
On a 20-page document with Qwen3.5-9B, this adds ${\sim}$8\% wall-clock latency; the dominant cost remains the LLM's autoregressive decoding.

\subsection{Boundary-Handling Ablation}
\label{sec:appendix_boundary_handling}

At the first and last document pages, \methodname uses self-fallback ($Z_0=Z_1$, $Z_{P+1}=Z_P$). We compare this convention with two alternatives in Table~\ref{tab:boundary_handling}. The blank-white-page variant replaces missing neighbors with blank white pages only at inference, using the checkpoint trained with self-fallback. The learned-null variant is trained through both Stage~A and Stage~B. The inference-only substitution measures sensitivity of the existing checkpoint; it does not compare strategies trained under identical conditions.

\begin{table}[t]
\caption{Boundary-handling comparison on MMLongBench-Doc and LongDocURL (\%).}
\label{tab:boundary_handling}
\centering
\small
\begin{tabular}{lcc}
\toprule
Boundary handling & MMLongBench-Doc & LongDocURL \\
\midrule
Self-fallback (ours) & 58.3 & 62.9 \\
Blank white page (inference only) & 58.1 & 62.7 \\
Learned null-page embedding (retrained) & 57.9 & 62.4 \\
\bottomrule
\end{tabular}
\end{table}

All variants are within 0.9 points. We retain self-fallback because it requires no additional trainable parameters. SVE~\citep{sve} also uses first-image self-fallback and compares it against a learned null embedding in its ablations.

\subsection{Gate Positional-Input Ablation}
\label{sec:appendix_gate_inputs}

Absolute row positions do not directly indicate where document content begins or ends when page margins vary. To assess the role of the gate positional features, we compare visual-token features alone, visual-token features with standard sinusoidal encoding of the absolute row index, and visual-token features with our content-frontier encoding. The remaining architecture, training data, objectives, and training schedule are unchanged.

\begin{table}[t]
\caption{Gate positional-input ablation. Benchmark scores are reported in \%. The final column is the ratio of mean gate activation on content tokens to that on margin tokens, using the ink-mask protocol in Appendix~\ref{sec:appendix_gate_aggregate}.}
\label{tab:gate_inputs}
\centering
\resizebox{\linewidth}{!}{%
\begin{tabular}{lccc}
\toprule
Gate input & MMLongBench-Doc & LongDocURL & Content/margin ratio \\
\midrule
Visual token only, no positional feature & 56.9 & 60.6 & 1.1$\times$ \\
Absolute-row sinusoidal encoding & 57.4 & 61.5 & 1.2$\times$ \\
Content-frontier encoding (ours) & 58.3 & 62.9 & 1.4$\times$ \\
\bottomrule
\end{tabular}%
}
\end{table}

Content-frontier encoding improves performance by 0.9 / 1.4 points over absolute-row sinusoidal encoding. Its content-to-margin activation ratio is 1.4$\times$, compared with 1.2$\times$ for absolute-row encoding and 1.1$\times$ without positional features.


\section{Re-pagination Protocol}
\label{sec:appendix_repagination}

We evaluate on the complete test sets of MMLongBench-Doc (1{,}082 samples) and LongDocURL (2{,}325 samples) with no exclusions.
Every page in every sample is split into two half-pages using the procedure in Algorithm~\ref{alg:repagination}.

\begin{algorithm}[h]
\caption{Controlled re-pagination}
\label{alg:repagination}
\begin{algorithmic}[1]
\REQUIRE Page image $I$ of height $H$, mode $\in \{\textsc{Preserve}, \textsc{Break}\}$
\ENSURE Two half-page images $I_{\mathrm{top}}, I_{\mathrm{bottom}}$
\STATE Compute row-wise ink density $\rho(y)$ and content bounding box
\IF{mode $=$ \textsc{Preserve}}
    \STATE Find all whitespace bands (contiguous rows with $\rho < 0.004$, height $\geq 8$\,px) in the middle 70\% of the page
    \STATE Select the widest band closest to $H/2$; cut at its midpoint $y^*$
\ELSIF{mode $=$ \textsc{Break}}
    \STATE Detect horizontal table rules (long dark runs, $\geq 3$ clustered)
    \IF{table detected}
        \STATE Cut at an \emph{internal} row rule nearest $H/2$
    \ELSE
        \STATE Find consecutive same-paragraph text lines (gap $\leq 1.8\times$ median spacing, horizontal overlap $\geq 45\%$)
        \STATE Cut between the pair nearest $H/2$
    \ENDIF
\ENDIF
\STATE $I_{\mathrm{top}} \leftarrow I[0 : y^*],\quad I_{\mathrm{bottom}} \leftarrow I[y^* : H]$
\end{algorithmic}
\end{algorithm}

Both modes produce exactly $2N$ half-pages from $N$ originals, preserving the total pixel content, NaViT resolution, and token allocation.
Questions and answers are unchanged; the only variable is the vertical position of the artificial boundary.
$\Delta_{\mathrm{pag}} = S_{\mathrm{preserve}} - S_{\mathrm{break}}$ isolates boundary-placement sensitivity by comparing two re-paginated variants with identical page counts.
$\Delta_{\mathrm{orig}} = S_{\mathrm{orig}} - S_{\mathrm{break}}$ additionally captures the effect of the increased page count.

\section{Aggregate Gate Activation Statistics}
\label{sec:appendix_gate_aggregate}

The qualitative gate visualizations in Section~\ref{sec:cross_page_analysis} (Figure~\ref{fig:gate_heatmaps}) show that \methodname's directional gates concentrate on boundary content and suppress blank margins.
To verify that this behavior is systematic rather than cherry-picked, we compute aggregate gate activation statistics over 200 randomly sampled triplets from the MMLongBench-Doc benchmark.

For each triplet, we record the sigmoid-activated gate values ($\sigma(s) \in [0,1]$) for all four directional gates and partition each page's tokens using an automatically computed ink mask (pixels darker than 95\% white = content).
We further divide each page into a \textbf{top band} (top 20\% of rows), \textbf{bottom band} (bottom 20\%), and \textbf{middle} region.

\paragraph{Content vs.\ margin selectivity.}
Across all gates and pages, mean activation on content regions is 0.27 vs.\ 0.19 on margins (1.4$\times$ ratio, Pearson $\rho = 0.25$ with the ink mask).
While this ratio is moderate, the gates are not designed to be binary content detectors; their primary role is to \emph{directionally} modulate cross-page attention flow, which the next analysis reveals.

\paragraph{Directional boundary specialization.}
The most striking pattern is that each gate learns to concentrate on the \emph{directionally appropriate} page boundary, with specialization learned under the combination of task loss and a weak geometric prior (Table~\ref{tab:gate_directional}).

\begin{table}[h]
\caption{Directional gate activation by page region, averaged over 200 document triplets (center page shown).
Gates interacting with the next page concentrate on the \textbf{bottom} band; gates interacting with the previous page concentrate on the \textbf{top} band.
The directional asymmetry (2.8-3.9$\times$) is observed with task supervision and a weak frontier prior ($\lambda = 10^{-3}$); their separate contributions are not isolated by these statistics.}
\label{tab:gate_directional}
\begin{center}
\small
\begin{tabular}{lccccc}
\toprule
Gate direction & \makecell{Top\\band} & Middle & \makecell{Bottom\\band} & \makecell{Relevant\\boundary} & \makecell{Relevant/\\Opposite} \\
\midrule
\multicolumn{6}{l}{\textit{Next-page gates (should concentrate on bottom)}} \\
send\_to\_next & 0.13 & 0.25 & \textbf{0.46} & bottom & 3.5$\times$ \\
recv\_from\_next & 0.16 & 0.26 & \textbf{0.45} & bottom & 2.8$\times$ \\
\midrule
\multicolumn{6}{l}{\textit{Previous-page gates (should concentrate on top)}} \\
recv\_from\_prev & \textbf{0.46} & 0.25 & 0.15 & top & 3.1$\times$ \\
send\_to\_prev & \textbf{0.47} & 0.24 & 0.12 & top & 3.9$\times$ \\
\bottomrule
\end{tabular}
\end{center}
\end{table}

Gates that mediate interaction with the \textbf{next} page (\texttt{send\_to\_next}, \texttt{recv\_from\_next}) activate 2.8-3.5$\times$ more strongly on the bottom boundary band than the top, precisely where content continues onto the next page.
Conversely, gates mediating interaction with the \textbf{previous} page (\texttt{recv\_from\_prev}, \texttt{send\_to\_prev}) activate 3.1-3.9$\times$ more on the top boundary band.
The middle region receives uniformly moderate activation (${\sim}$0.25) across all gates.

This directional specialization is guided by the Stage~A frontier prior (Appendix~\ref{sec:gate_details}), which pairs each gate with its directionally appropriate ink frontier (top frontier for previous-page gates, bottom frontier for next-page gates, $\lambda_{\mathrm{frontier}} = 10^{-3}$).
The aggregate statistics show consistent directional gate patterns under this combination of geometric and task supervision.
The pattern is consistent across all three page positions in the triplet (previous, center, next); center-page values are shown in Table~\ref{tab:gate_directional}, with the full breakdown available in the released statistics file.

These aggregate measurements complement the qualitative visualizations in the main paper and provide quantitative evidence that the learned gate behavior is a systematic, directionally-specialized property of the module, not an artifact of selected examples.

\section{Analysis of Post-Encoder Representation Leakage}
\label{sec:appendix_representation_leakage}

As noted in Section~\ref{sec:training}, a natural candidate for pre-training \methodname in Stage~A is self-supervised \emph{boundary reconstruction}, which masks visual tokens near page margins and trains the module to reconstruct them from neighboring pages. However, this objective performs poorly at the \emph{post-encoder} insertion point, consistent with a same-page shortcut that we refer to as \textbf{post-encoder representation leakage}.

\begin{figure}[h]
\centering
\includegraphics[width=0.6\textwidth]{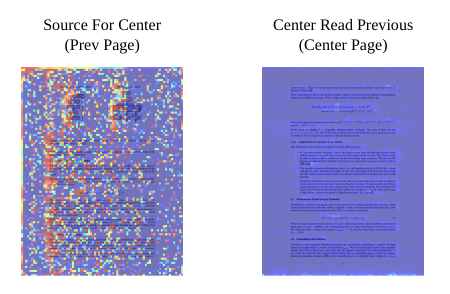}
\caption{\textbf{Receive gates under post-encoder boundary reconstruction.}
When trained with self-supervised boundary reconstruction post-encoder, the receive gates on the center page (\emph{right}) fall near zero (dark blue), strongly downweighting cross-page attention even though the source features on the previous page (\emph{left}) contain valid content signals. This behavior is consistent with a same-page shortcut: global self-attention in the vision encoder can make information from interior tokens available in the boundary representations.}
\label{fig:gate_collapse}
\end{figure}

\paragraph{Mechanism.}
Modern VLM vision towers (such as NaViT or ViT backbones) employ global intra-page self-attention across all visual patches of a single page image. Consequently, by the time visual representations reach the post-encoder stage, tokens at the top or bottom boundaries of page~$p$ have already absorbed rich global context from unmasked interior tokens on the same page. 

When a self-supervised reconstruction loss is applied on post-encoder visual features:
\begin{enumerate}[leftmargin=*]
    \item Unmasked features from the \emph{same} page may provide information useful for predicting masked boundary features.
    \item The module can consequently reduce its reliance on neighbor pages, even when those pages contain continuation evidence.
    \item In the observed example, the receive gates approach zero (Figure~\ref{fig:gate_collapse}), strongly downweighting cross-page attention. This behavior is consistent with reliance on a same-page shortcut.
\end{enumerate}

The observed collapse suggests that this post-encoder reconstruction setup admits a same-page shortcut. It motivates supervised Stage~A tasks whose target answers require cross-page evidence, but does not rule out other self-supervised objectives or masking strategies.

\section{Extended Layer-Wise Attention Trace Analysis}
\label{sec:appendix_extended_attention_trace}

\begin{figure}[h]
\centering
\includegraphics[width=0.95\textwidth]{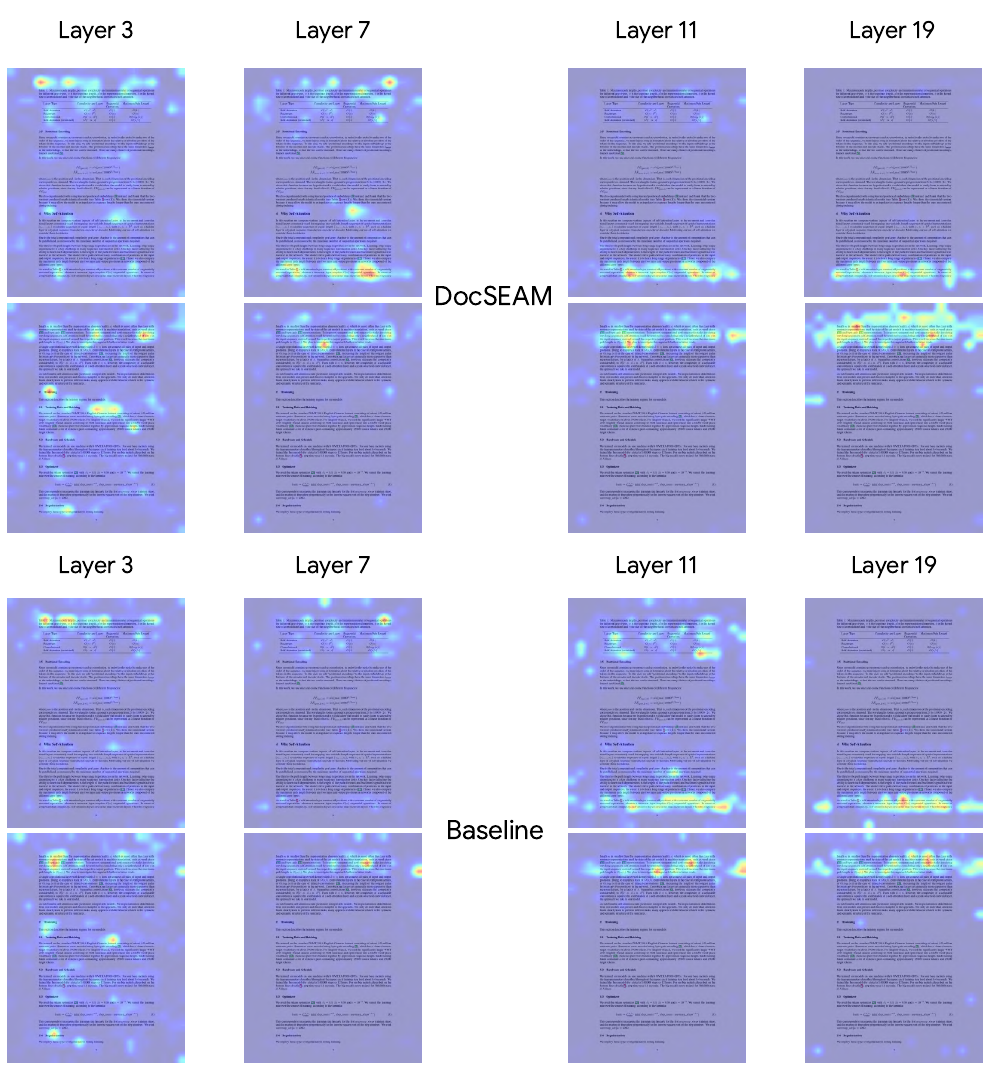}
\caption{\textbf{Extended layer-wise attention trace across deep LLM layers.}
Comparison of attention maps between \methodname (top) and the stateless baseline (bottom) at Layers 3, 7, 11, and 19 for a cross-page span localization query.
With \methodname, attention sharply concentrates on the boundary-spanning ground-truth evidence as early as Layer~7 and maintains a sharp, focused pattern through Layer~19.
In contrast, the baseline exhibits diffuse attention at shallow layers (Layers 3 and 7), only forms a relevant pattern around Layer~11, and retains noticeable background noise even at Layer~19.}
\label{fig:extended_attention_trace}
\end{figure}

To complement the early-layer attention trace presented in Section~\ref{sec:cross_page_analysis}, we visualize the progression of LLM layer-wise attention maps across both shallow and deep layers (Layers 3, 7, 11, and 19) in Figure~\ref{fig:extended_attention_trace}.

\paragraph{Observations:}
\begin{enumerate}[leftmargin=*]
    \item \textbf{Early Convergence (Layers 3-7):} At Layer~3, both models show initial exploratory attention across the page body. By Layer~7, \methodname already exhibits strong, localized activation on the target text span across the page boundary, whereas the baseline remains scattered without a clear focal point.
    \item \textbf{Sustained Focus at Deep Layers (Layers 11-19):} In deeper layers (Layer~11 and Layer~19), \methodname maintains a high signal-to-noise ratio, keeping attention tightly bounded to the cross-page evidence required for answer generation. While the baseline eventually locates the region at Layer~11, its attention map at Layer~19 displays residual activation across irrelevant text regions, indicating a more diffuse attention pattern in this example.
\end{enumerate}

\section{Additional Directional Salience Case Studies}
\label{sec:additional_visualizations}

To further evaluate the generalization of \methodname across diverse document layouts and directly address potential concerns regarding multi-column formatting, we present additional qualitative case studies illustrating the directional salience gating mechanism.

\begin{figure}[h]
\centering
\includegraphics[width=\textwidth]{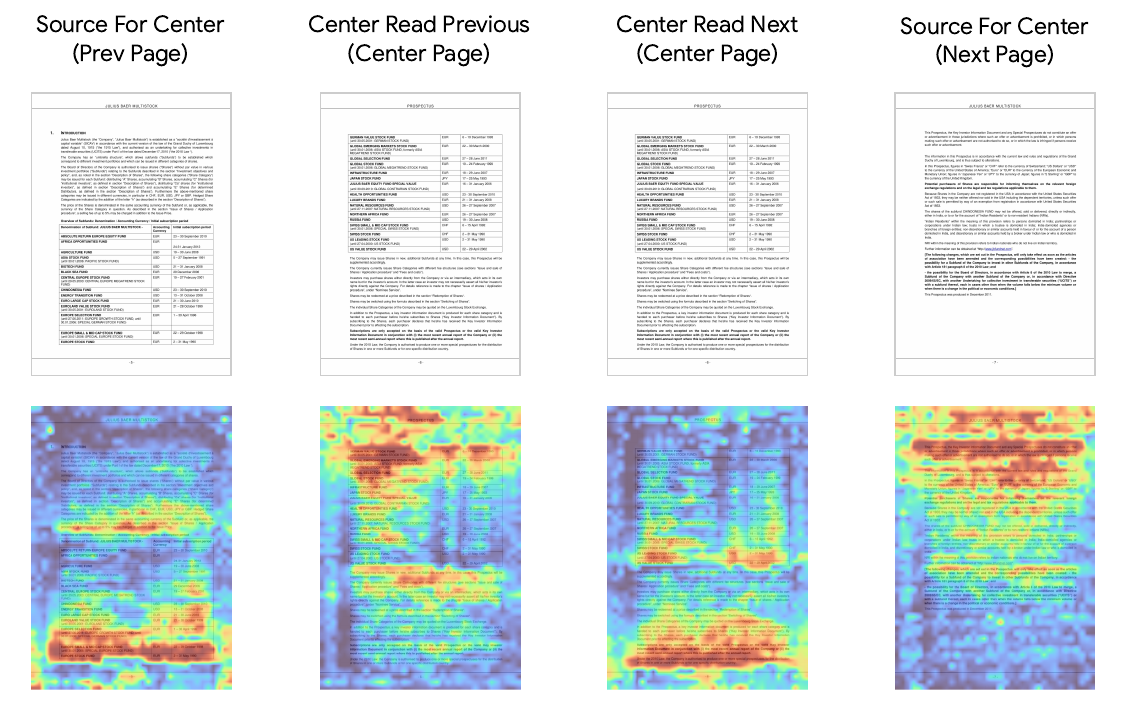}
\caption{\textbf{Case study of directional salience gate activation across a table continuation.} Attention heatmaps across three consecutive single-column document pages containing a multi-page table split across a page boundary. \methodname dynamically activates cross-page attention to connect table rows truncated at the bottom of the previous page with their continuation at the top of the center page, while suppressing uninformative margins.}
\label{fig:second_heat_map}
\end{figure}

\begin{figure}[h]
\centering
\includegraphics[width=\textwidth]{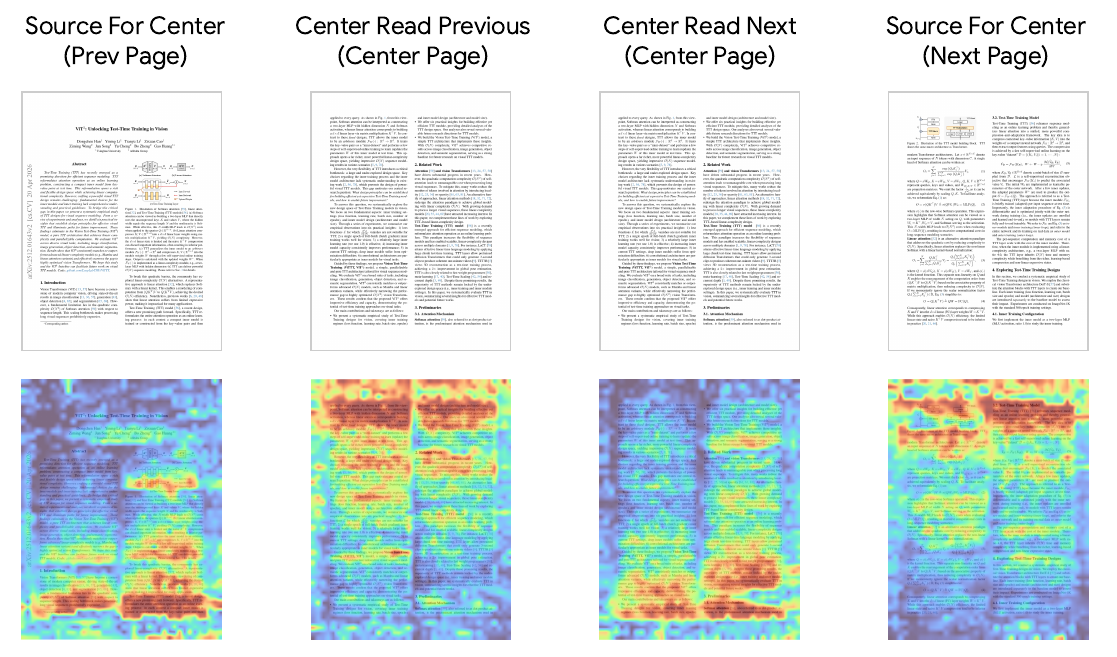}
\caption{\textbf{Generalization to complex two-column layouts and robustness to structural interruptions.} Attention routing across consecutive two-column scientific paper pages. The displayed activation patterns overlap text columns and continuation regions despite intervening figures.}
\label{fig:twocol_heat_map}
\end{figure}

\paragraph{Key Observations across Layout Types:}
\begin{enumerate}
    \item \textbf{Cross-Boundary Table Alignment (Single-Column, Figure~\ref{fig:second_heat_map}):} The previous page (\emph{Source For Center}, left) ends with an incomplete table detailing financial fund allocations. As shown in the \emph{Center Read Previous} panel, when the center page attends backward, attention activates intensely (highlighted in bright red/yellow) over the table continuation at the top of the center page, consistent with the intended boundary-routing behavior.

    \item \textbf{Generalization to Two-Column Layouts (Figure~\ref{fig:twocol_heat_map}):} When deployed on dense, two-column scientific documents, \methodname demonstrates layout-aware gating behavior. In the \emph{Center Read Previous} panel, attention isolates the two text columns while the inter-column gutter receives near-zero activation, indicating that the gates distinguish column regions from non-content space without explicit column supervision. Activation is elevated across both column regions, which is expected in a two-column layout: the right column typically continues onto the next page, while the left column may reference figures, tables, or equations that appear on adjacent pages, making both columns potential sites of cross-page information flow.

    \item \textbf{Robustness to Intervening Visual Elements (Figure~\ref{fig:twocol_heat_map}):} Similar to the single-column case in the main paper (where Table~1 intervenes in \emph{Attention Is All You Need}), the next page contains a large figure at the top with continuation text below it. In the \emph{Center Read Next} and \emph{Source For Center (Next Page)} panels, \methodname produces two distinct bright regions on this page: one on the figure and one on the paragraph continuation beneath it, with intermediate space receiving lower activation. A purely positional gate would produce monotonically decreasing activation from the top of the page; instead, the gate selectively identifies both the figure and the text continuation as relevant content regions.

    \item \textbf{Selective Margin \& Header Suppression:} In the displayed single- and multi-column examples, blank side margins, running headers, and page numbers receive low attention (dark blue).
\end{enumerate}

\subsection{Quantitative Column-Asymmetry Analysis on Two-Column Documents}
\label{sec:appendix_twocol_asymmetry}

The qualitative observations above suggest that \methodname's gates respect two-column layout structure.
To verify this quantitatively, we measure the directional asymmetry of gate activation across left and right columns.

In a standard two-column document, the reading order flows from the left column to the right column on each page, then continues to the left column of the next page.
Consequently, the \emph{continuation-relevant} column differs by gate direction:
gates mediating interaction with the \textbf{next} page (\texttt{send\_to\_next}, \texttt{recv\_from\_next}) should concentrate on the \textbf{right} column (where text exits the page), while gates mediating interaction with the \textbf{previous} page (\texttt{recv\_from\_prev}, \texttt{send\_to\_prev}) should concentrate on the \textbf{left} column (where text enters from the preceding page).

\begin{figure}[h]
\centering
\includegraphics[width=\textwidth]{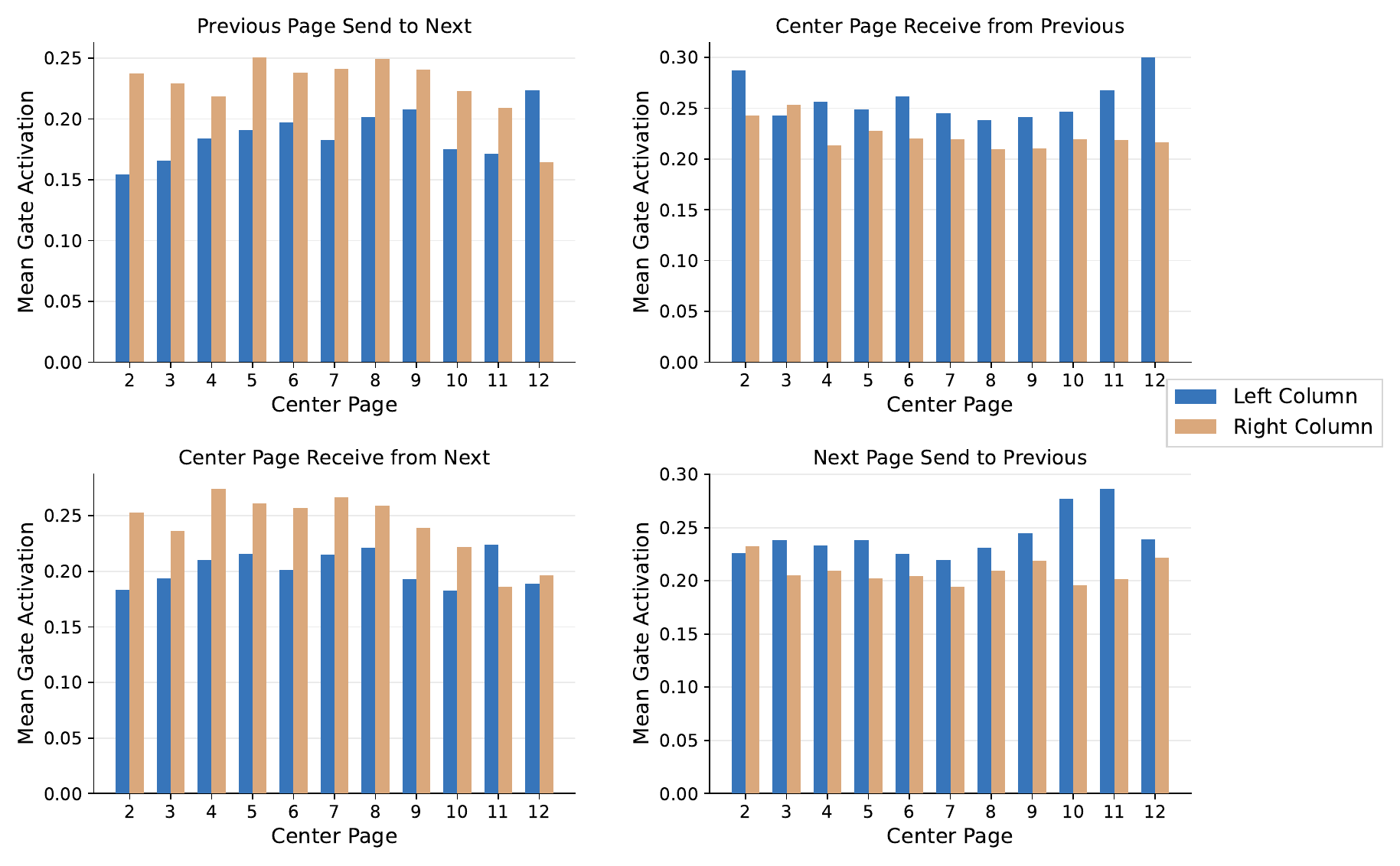}
\caption{\textbf{Per-triplet left vs.\ right column gate activation on a two-column document.}
Each subplot shows one directional gate across 11 consecutive-page triplets. The continuation-relevant column (right for next-page gates, left for previous-page gates) shows consistently higher activation in 10 out of 11 triplets.}
\label{fig:twocol_per_triplet}
\end{figure}

\paragraph{Per-triplet column asymmetry.}
We extract gate activations for all 11 consecutive-page triplets of the same 13-page two-column scientific paper we showed earlier and split each gate heatmap into left and right halves (excluding a 5\% center gutter).
Figure~\ref{fig:twocol_per_triplet} shows the per-triplet comparison: across all four gates, the continuation-relevant column shows higher mean activation in \textbf{10 out of 11 triplets} (91\%).

\begin{figure}[h]
\centering
\includegraphics[width=\textwidth]{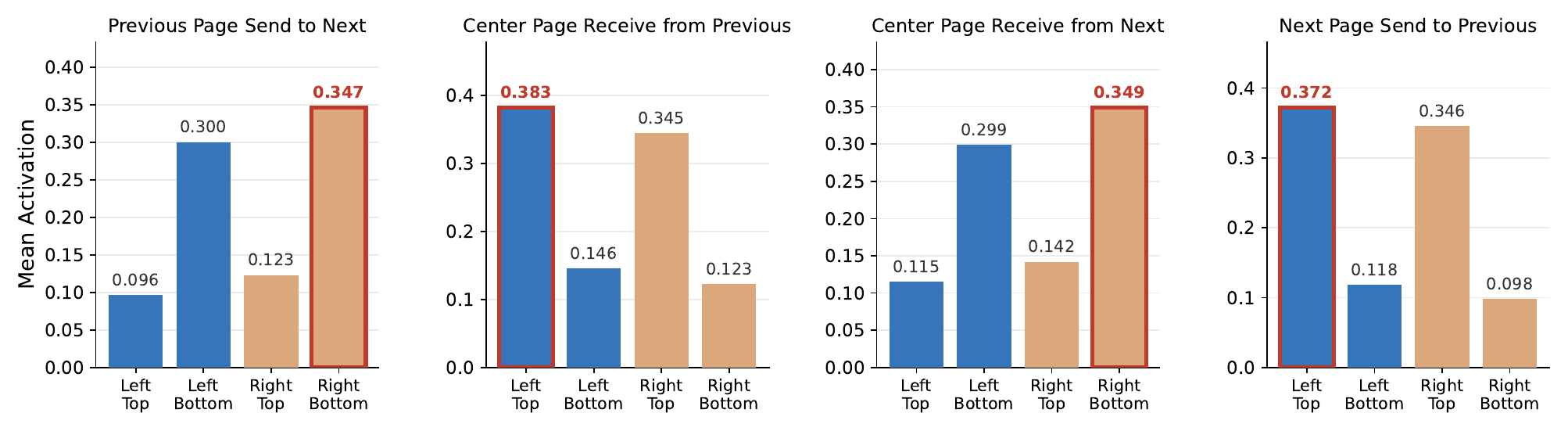}
\caption{\textbf{Quadrant-level gate activation averaged over all triplets.}
Each page is divided into four quadrants: Left-Top and Left-Bottom (left column, blue), Right-Top and Right-Bottom (right column, orange). The top and bottom bands each span 30\% of the page height.
Red-outlined bars mark the continuation-relevant corner for each gate direction (right-bottom for next-page gates, left-top for previous-page gates).
The continuation-relevant corner activates 3.0 to 3.8$\times$ more than the diagonally opposite corner, matching the two-column reading order.}
\label{fig:twocol_quadrant}
\end{figure}

\paragraph{Quadrant-level analysis.}
To further localize the activation pattern, we divide each page into four quadrants (left-top, left-bottom, right-top, right-bottom using a 30\% boundary band) and average across all triplets.
Figure~\ref{fig:twocol_quadrant} shows the result.
The continuation-relevant \emph{corner} activates 3.0 to 3.8$\times$ more than the diagonally opposite corner:
\texttt{send\_to\_next} concentrates on right-bottom (0.347 vs.\ 0.096 at left-top, 3.6$\times$);
\texttt{recv\_from\_prev} on left-top (0.383 vs.\ 0.123 at right-bottom, 3.1$\times$);
\texttt{recv\_from\_next} on right-bottom (0.349 vs.\ 0.115 at left-top, 3.0$\times$);
\texttt{send\_to\_prev} on left-top (0.372 vs.\ 0.098 at right-bottom, 3.8$\times$).
This pattern matches the expected two-column reading order: text exits a page from the right-column bottom and enters the next page at the left-column top.

\paragraph{Multi-document validation.}
To confirm that these patterns are not specific to a single document, we repeat the analysis across 30 two-column documents sampled from the MMLongBench-Doc and LongDocURL test set, yielding a total of 247 consecutive-page triplets.
Averaged across all documents, the continuation-relevant column dominates in 94\% of triplets, and the continuation-relevant corner activates  2.6 to 3.2$\times$ more than the opposite corner.
These aggregate statistics are consistent with the single-document results reported above and confirm that \methodname's column-aware gating behavior generalizes across documents.

\section{Qualitative Answer Comparisons}
\label{sec:appendix_answer_comparisons}

\begin{figure}[t]
\centering
\includegraphics[width=\textwidth]{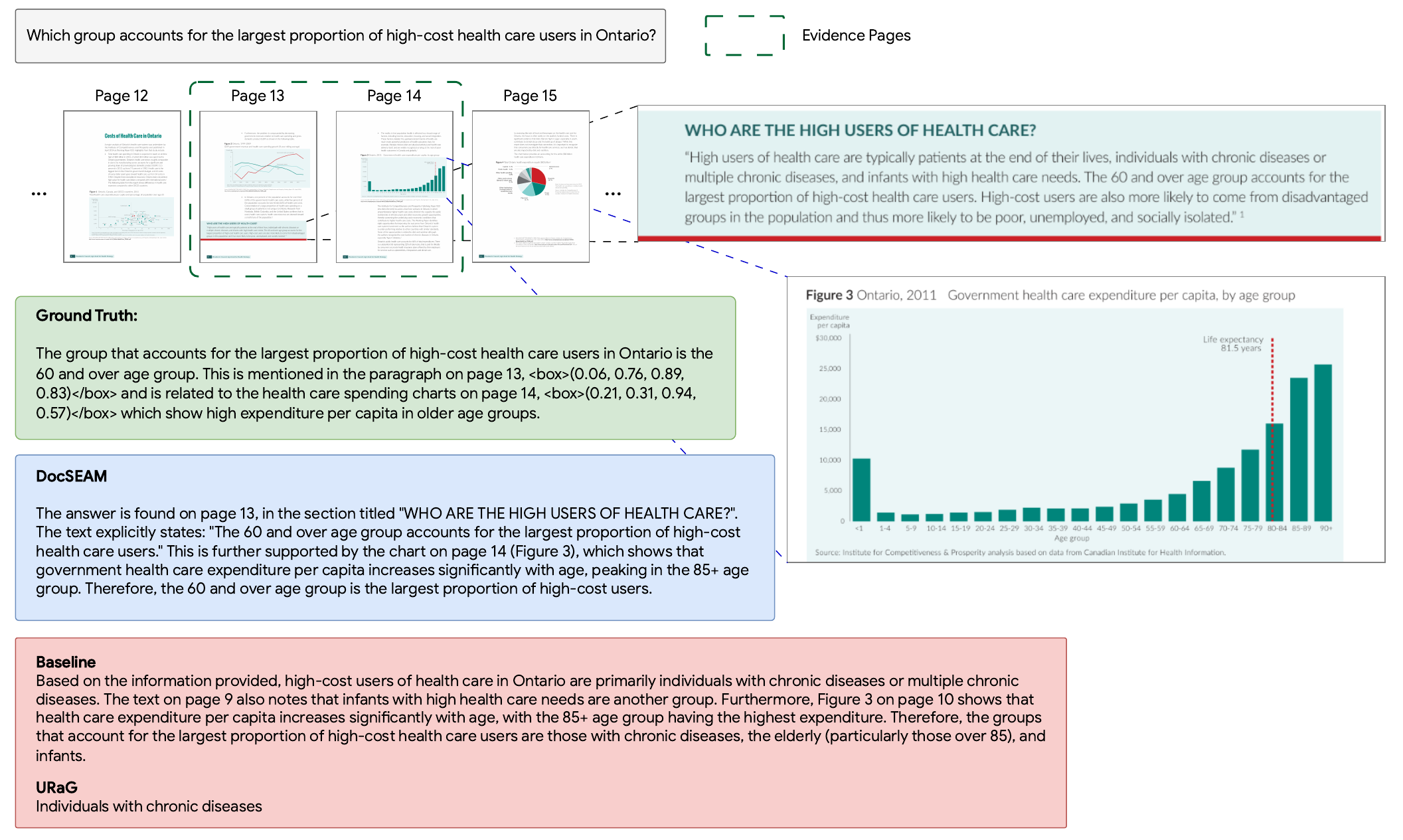}
\caption{\textbf{Qualitative comparison on multi-page cross-modal QA.} Given a question spanning a text description (Page 13) and a supporting chart (Page 14), \methodname correctly identifies the target demographic group with precise page attribution, whereas the baseline hallucinates page citations and URaG produces an ungrounded fragment.}
\label{fig:qualitative_health}
\end{figure}

\begin{figure}[t]
\centering
\includegraphics[width=\textwidth]{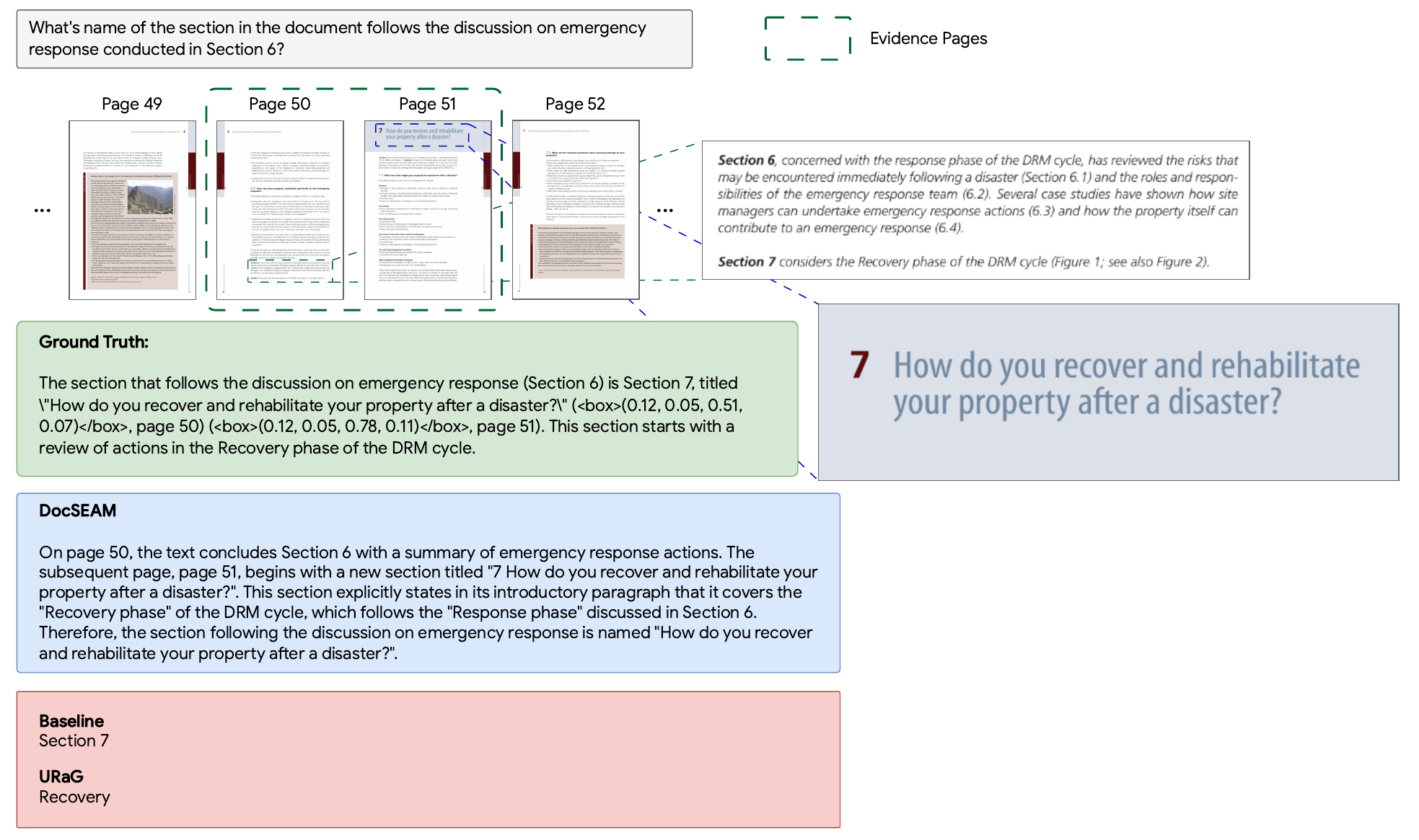}
\caption{\textbf{Qualitative comparison on section tracking across page breaks.} When identifying a section title continuing over a page boundary (Pages 50-51), \methodname extracts the complete section title and context, while baseline models fail to retrieve the full title.}
\label{fig:qualitative_section}
\end{figure}

To visually inspect how \methodname improves multi-page document reasoning over baseline models, we examine representative qualitative examples where answering the question requires integrating information across adjacent pages. Figures~\ref{fig:qualitative_health} and~\ref{fig:qualitative_section} present side-by-side comparisons of predictions from \methodname, the matched baseline, and URaG, alongside ground-truth annotations and document evidence.

\paragraph{Case 1: Cross-Page \& Cross-Modal Evidence Integration (Figure~\ref{fig:qualitative_health}).}
In this example from a multi-page health report, the question asks which demographic group accounts for the largest proportion of high-cost health care users.
The relevant text appears on Page~13, while a supporting chart is located on Page~14.
\methodname predicts the correct demographic group (\emph{``60 and over age group''}) and grounds its explanation with page references.
In contrast, the stateless baseline cites incorrect pages (Page~9 and Page~10 instead of Page~13 and Page~14) and lists multiple risk groups; these outputs alone do not identify the internal cause of the error.
URaG outputs a brief fragment (\emph{``Individuals with chronic diseases''}) without citing evidence.

\paragraph{Case 2: Structural Tracking Across Section Boundaries (Figure~\ref{fig:qualitative_section}).}
This example evaluates structural section tracking over document page breaks. The prompt asks for the title of the section following Section 6 (\emph{``Emergency Response"}). Section 6 concludes at the bottom of Page~50, and Section 7 begins at the top of Page~51. 
\methodname retrieves the complete section title (\emph{"7 How do you recover and rehabilitate your property after a disaster?"}). Conversely, the baseline model only produces a generic label (\emph{"Section 7"}), and URaG returns only a single word (\emph{"Recovery"}), missing the complete heading required by the ground truth.

\section{Re-pagination Training Augmentation Control}
\label{sec:appendix_repagination_augmentation}

To test whether exposure to re-paginated documents reduces the baseline's pagination sensitivity, we augment the SFT baseline's Stage~B data with re-paginated training documents, using equal proportions of structure-preserving and structure-breaking cuts. The remaining training settings are unchanged. Table~\ref{tab:repagination_augmentation} evaluates both original and re-paginated inputs using the protocol in Appendix~\ref{sec:appendix_repagination}.

\begin{table}[t]
\caption{SFT with re-pagination augmentation. Orig., Pres., and Break denote original, structure-preserving, and structure-breaking inputs, respectively. Scores are accuracy (\%).}
\label{tab:repagination_augmentation}
\centering
\resizebox{\linewidth}{!}{%
\begin{tabular}{lcccccc}
\toprule
\multirow{2}{*}{Model} & \multicolumn{3}{c}{MMLongBench-Doc} & \multicolumn{3}{c}{LongDocURL} \\
\cmidrule(lr){2-4}\cmidrule(lr){5-7}
 & Orig. & Pres. & Break & Orig. & Pres. & Break \\
\midrule
SFT & 44.5 & 42.1 & 40.9 & 42.8 & 41.5 & 39.9 \\
SFT + re-pagination augmentation & 44.4 & 41.9 & 40.7 & 43.5 & 42.1 & 40.3 \\
\methodname & 58.3 & 57.6 & 56.8 & 62.9 & 61.8 & 61.1 \\
\bottomrule
\end{tabular}%
}
\end{table}

Augmentation has little effect on MMLongBench-Doc and yields small gains on LongDocURL (+0.7 on original, +0.6 on preserved, and +0.4 on broken inputs). The pagination penalty remains 1.2 on MMLongBench-Doc and changes from 1.6 to 1.8 on LongDocURL. Thus, this configuration does not reduce boundary-placement sensitivity. \methodname exceeds the augmented baseline by 13.9 / 19.4 points on original inputs and 16.1 / 20.8 points on broken inputs. Under this augmentation protocol, a substantial gap remains on both original and re-paginated inputs.

\section{Comparison with Stateful Visual Encoders}
\label{sec:appendix_sve}

We adapt the released SVE~\citep{sve} implementation to Qwen3.5-9B. Following its single-stage fine-tuning setup, we train it on the combined Stage~A and Stage~B data, as for the stateless SFT baseline. SVE inserts Cross+FFN before every vision-encoder block, with cloned input projections, zero-initialized output projections, stop-gradient on predecessor features, and first-page self-fallback. Table~\ref{tab:sve_comparison} also includes the zero-shot baseline and the post-encoder full-triplet-attention variant without a mask or gates from Table~\ref{tab:ablations}.

\begin{table}[t]
\caption{Comparison with SVE on Qwen3.5-9B (accuracy, \%). LongDocURL is additionally evaluated on adjacent two-page and non-adjacent-only evidence subsets, as defined in Appendix~\ref{sec:appendix_stratified}.}
\label{tab:sve_comparison}
\centering
\small
\resizebox{\linewidth}{!}{%
\begin{tabular}{lcccc}
\toprule
\multirow{2}{*}{Method} & \multirow{2}{*}{\makecell{MMLong\\Bench-Doc}} & \multicolumn{3}{c}{LongDocURL} \\
\cmidrule(lr){3-5}
 & & Overall & \makecell{Adjacent\\two-page} & \makecell{Non-adjacent\\only} \\
\midrule
Zero-shot baseline & 57.1 & 54.5 & 58.0 & 48.8 \\
SFT baseline & 44.5 & 42.8 & 47.0 & 40.0 \\
SVE (in-encoder, causal) & 45.2 & 43.4 & 47.6 & 38.7 \\
Post-encoder, bidirectional, no mask or gates & 55.2 & 57.8 & 64.1 & 50.3 \\
\methodname & 58.3 & 62.9 & 69.5 & 54.7 \\
\bottomrule
\end{tabular}%
}
\end{table}

SVE yields small aggregate gains over SFT (+0.7 / +0.6 points), whereas \methodname yields larger gains (+13.8 / +20.1). Although SVE directly accesses only the preceding page at each layer, information can propagate from earlier pages through successive encoder layers. On the non-adjacent-only subset, \methodname achieves 54.7\%, compared with 38.7\% for SVE and 40.0\% for SFT. Thus, SVE's potential for longer-range propagation does not translate into an advantage on this subset. \methodname enriches local representations while leaving integration across distant evidence pages to the LLM.

\begin{figure}[t]
\centering
\includegraphics[width=\textwidth]{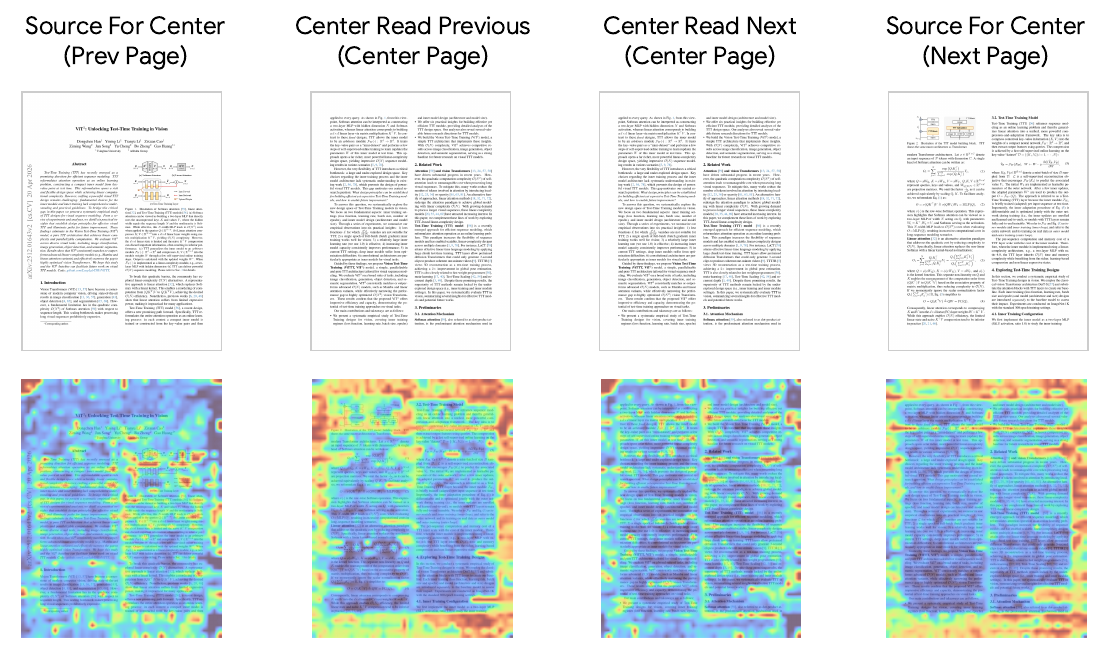}
\caption{\textbf{Causal SVE cross-page attention and update magnitudes.} Source panels show attention averaged over query tokens and heads, relative to uniform attention. Receiving panels show the L2 norm of projected cross-attention updates, including projection bias. Each panel is independently scaled for display.}
\label{fig:sve_visualization}
\end{figure}

SVE's original evaluations emphasize visual comparison and change detection, whereas document QA requires precise reading and cross-page linking. In Figure~\ref{fig:sve_visualization}, source attention is diffuse, and large projected-update norms also occur in blank margins. This example suggests limited specialization for boundary-continuation evidence, consistent with SVE's small aggregate gains.

\end{document}